\documentclass{article} 
\usepackage[final]{colm2026_conference}

\usepackage{hyperref}
\usepackage{url}
\usepackage{microtype}
\usepackage{booktabs}
\usepackage{pgfplots}
\usepackage{array}
\usepackage{longtable}
\usepackage{xcolor}
\usepackage{xspace}
\usepackage{subcaption}
\usepackage{pifont}
\usepackage{pgfplotstable}
\usepgfplotslibrary{groupplots}
\usepackage{wrapfig}
\usepackage{tcolorbox}
\pgfplotsset{compat=1.18}
\definecolor{cbBlue}{RGB}{0,114,178}
\definecolor{cbOrange}{RGB}{230,159,0}
\definecolor{cbGreen}{RGB}{0,158,115}
\definecolor{cbYellow}{RGB}{240,228,66}
\definecolor{cbDarkYellow}{RGB}{144,137,40}
\definecolor{cbRed}{RGB}{204,121,167}
\definecolor{cbPurple}{RGB}{86,180,233}
\definecolor{cbGray}{RGB}{128,128,128}
\definecolor{cbBlack}{RGB}{0,0,0}
\definecolor{maroon}{RGB}{85,0,0}

\usepackage{color-edits}
\addauthor{gn}{magenta}

\newcommand{\ap}[1]{}
\newcommand{\ours}{\textsc{Oolong}\xspace}

\newcommand{\cmark}{\textcolor{green!60!black}{\ding{51}}}  
\newcommand{\xmark}{\textcolor{red}{\ding{55}}}              
\newcommand{\smark}{\textcolor{orange!80!yellow}{\ding{61}}} 

\usepackage{lineno}

\definecolor{darkblue}{rgb}{0, 0, 0.5}
\hypersetup{colorlinks=true, citecolor=darkblue, linkcolor=darkblue, urlcolor=darkblue}

\title{\ours: Evaluating Long Context \\Reasoning and Aggregation Capabilities}

\author{Amanda Bertsch \qquad
Adithya Pratapa\thanks{Work conducted while the author was a student at Carnegie Mellon University.} \qquad \\
\textbf{Teruko Mitamura} \qquad
\textbf{Graham Neubig} \qquad \textbf{Matthew R. Gormley }\qquad \\[0.5em]
Carnegie Mellon University \\
\texttt{\{abertsch,vpratapa,teruko,gneubig,mgormley\}@cs.cmu.edu}\\
\href{https://github.com/abertsch72/oolong}{\includegraphics[height=0.4cm]{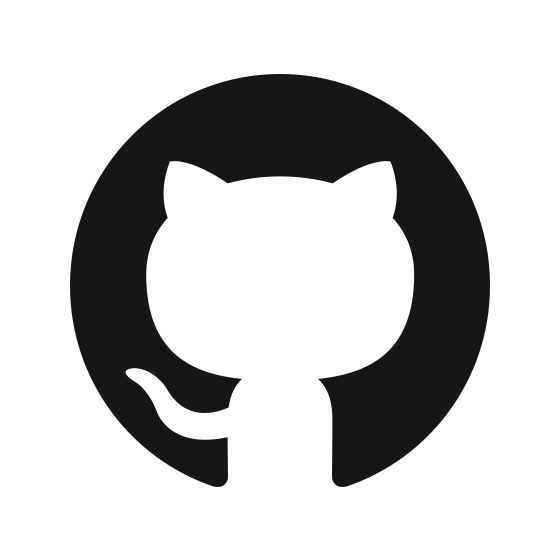} \textbf{abertsch72/oolong}} ~ ~ ~ \href{https://huggingface.co/datasets/oolongbench}{\includegraphics[height=0.4cm]{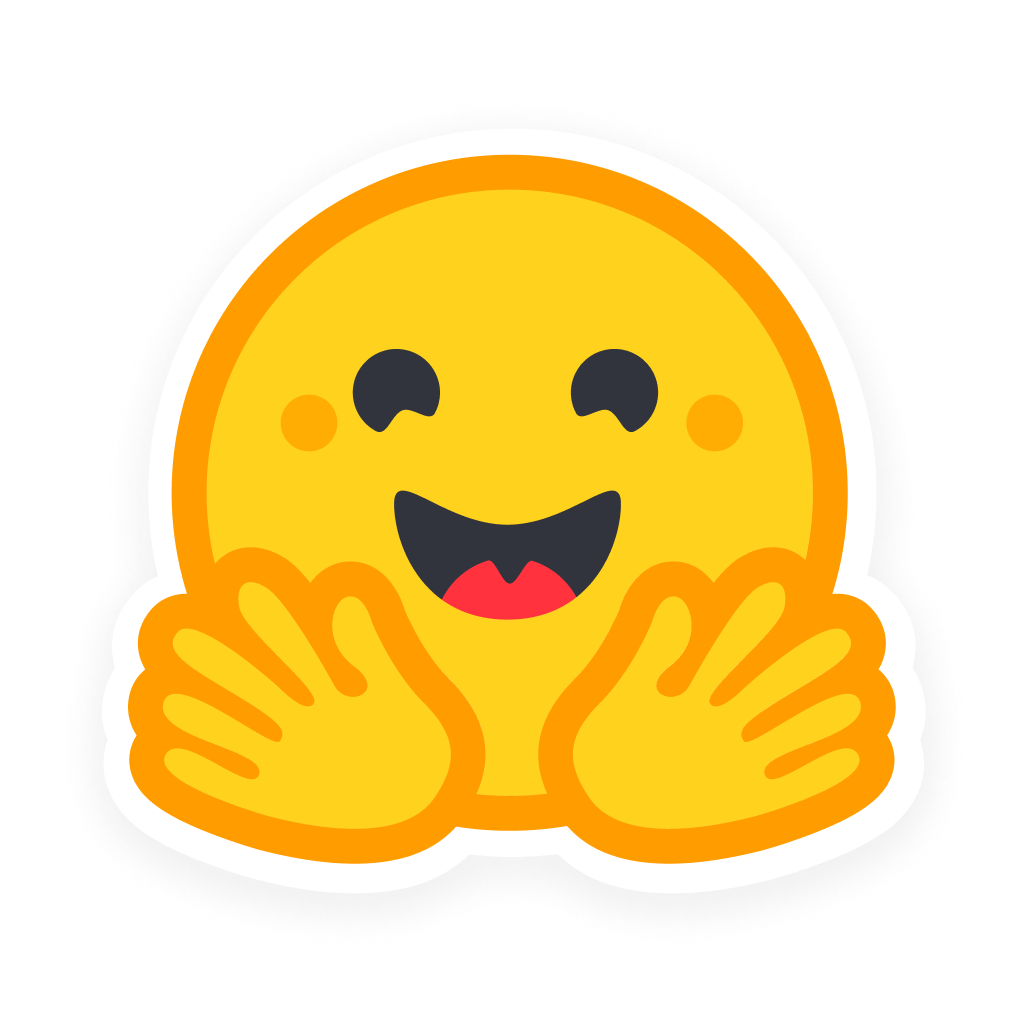} \textbf{oolongbench/oolong-[synth, real]}}
}

\begin{document}

\ifcolmsubmission
\linenumbers
\fi

\maketitle

\begin{abstract}
    Many important applications of long-context models require reasoning over large collections of items—summarizing trends across thousands of customer interactions, identifying patterns in research corpora, or analyzing behavioral data at scale. Evaluating whether models can analyze items atomically and aggregate findings requires information-dense tasks that make use of the full context, like writing a summary or describing trends across a collection. Yet these types of tasks often result in long-form outputs which are challenging to evaluate; as a result, current long-context benchmarks predominantly test retrieval-style tasks, where the correct answer depends on a small fraction of tokens and the rest is noise. We introduce \ours, a benchmark of distributional reasoning tasks that require analyzing individual chunks of text atomically and aggregating results. \ours includes synthetic tasks (\ours-synth) that enables controlled ablation of reasoning components, and a downstream setting (\ours-real) that requires reasoning over realistic conversational data. Tasks involve classification, counting, filtering by metadata, and temporal reasoning. Frontier models struggle at moderate context lengths at this task, suggesting \ours captures capabilities not yet addressed by scale alone. We release the data and evaluation harness for \ours to enable further development of models that can reason over large quantities of text.

\end{abstract}

\section{Introduction}

In the last several years, the exponentially increasing context lengths of LLMs have enabled many new applications, including reasoning models \citep{guo2025deepseek}, many-shot prompting \citep{bertsch2024context}, and repository-level code generation \citep{jimenez2023swe}. However, despite the rapid improvement in this area, 
 long-context aggregation remains a challenging capability to study. Tasks like summarization are realistic but produce long and subjective outputs that are challenging to evaluate. Other tasks require only a subset of the input, which could be extracted from the long context before solving (e.g. using a RAG system).

Multi-step reasoning datasets make progress towards measuring this capability, but often lack realism – inputs may be artificial (e.g. random repeated words in RULER \citep{hsieh-2024-ruler}), the posed questions may be unrealistic (e.g. “identify the second poem about this topic” in MRCR \citep{openai-2025-mrcr}), and the relevant context may be embedded in distractor text from a different domain or a well-known text (e.g. small state updates embedded in Project Gutenberg novels in BABILong \citep{kuratov2024babilongtestinglimitsllms}). We argue that aggregation is best measured in realistic settings with multiple \textit{types} of aggregative component tasks-- including the ability to draw inferences across different sections of input and perform basic numeracy operations with this information. 
We further discuss these desiderata in \autoref{tab:related_work_comp}.

\begin{figure*}[t]
    \centering
    \includegraphics[width=0.8\linewidth]{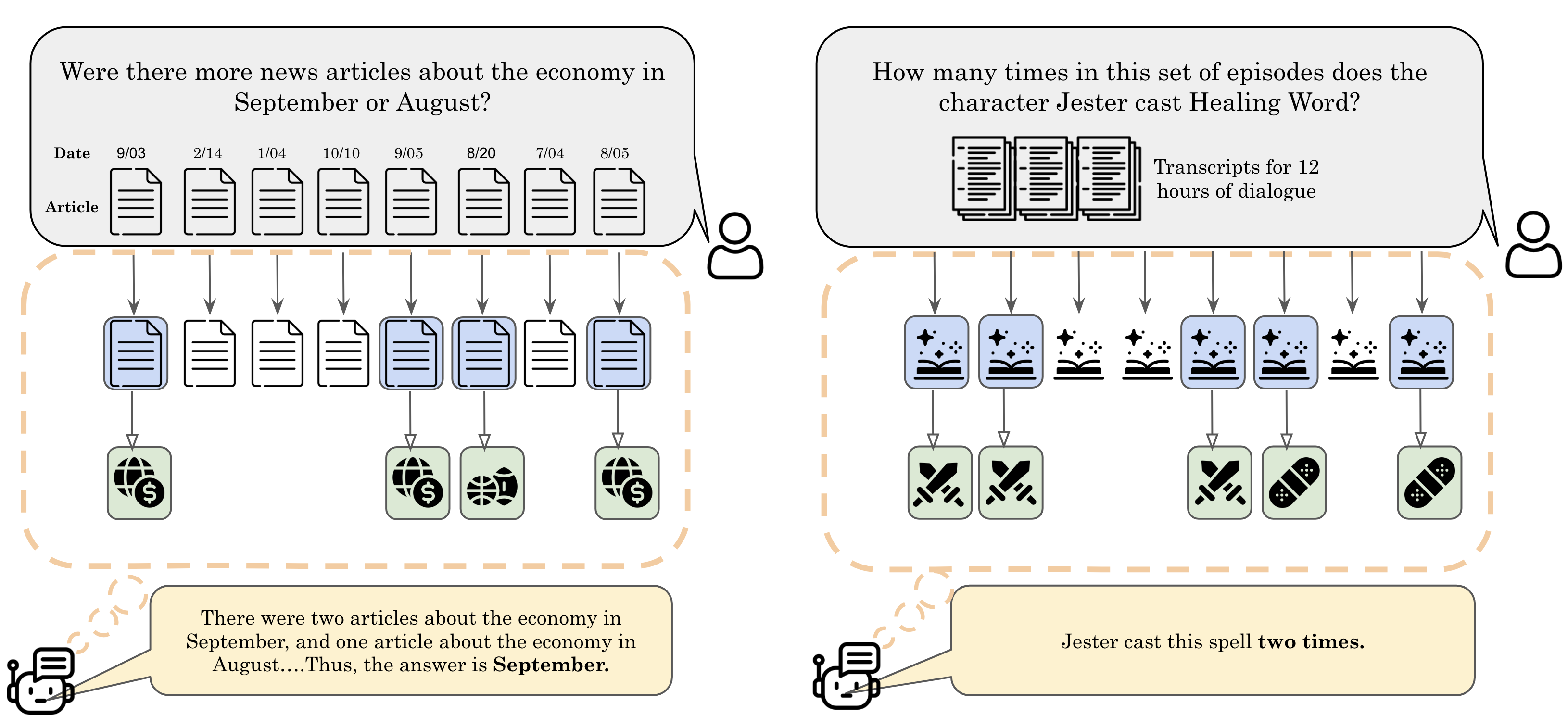}
    \caption{\ours poses questions that require performing a multi-step information aggregation process to determine the solution. \ours-synth uses ICL-based tasks, which could be easily decomposed and solved iteratively, as a proxy for real-world aggregation tasks over long inputs. \ours-real poses challenging information aggregation questions over transcripts from live-action Dungeons \& Dragons shows, which can not be easily decomposed into component pieces.}
    \label{fig:figure1}
    \vspace{-8pt}
\end{figure*}

We propose \ours, a benchmark that requires multi-hop reasoning over long inputs to produce easily verifiable outputs.
\ours is separated into two task sets, presented in \autoref{fig:figure1}. \ours-synth (\S \ref{sec:synth}) is a set of naturalistic synthetic tasks constructed from existing in-context learning datasets. These tasks require implicitly labeling the examples in-context to reason over distributional properties of the labels (\textit{counting tasks}), over user-specific patterns (\textit{user tasks}), and over changes in label distributions over time (\textit{temporal tasks}). \ours-real (\S \ref{sec:real}) poses these types of questions over real data that is not so easily separable into component parts-- asking challenging questions about character actions 
from live-play Dungeons \& Dragons transcripts, using human-annotated gold answers. 

Each task requires identifying the relevant segments of the input; this ranges from only a handful of instances or lines of dialogue to questions that require the use of \emph{every} line of the input. The relevant segments must be classified or categorized, and these individual decisions must be aggregated to produce a final answer.
By framing the benchmark around problems that are \emph{simple individually}, \ours measures capability on long context reasoning and not accuracy on the underlying task. By requiring identification of relevant context, classification decisions over that context, and numeracy skills to produce statistics about the input, \ours requires models to perform multi-step reasoning over long inputs. In \autoref{tab:related_work_comp}, we compare \ours with most similar long-context reasoning benchmarks.


Although these component capabilities have been well studied and frontier models demonstrate strong performance in each (\citet{yen2024helmet}, \citet{agarwal2024manyshotincontextlearning}, \textit{inter alia}), we find that models struggle with information aggregation as the input length grows. Most models score below 50\% on \ours at 128K context. 
We study why this task is so challenging 
and find that identification and aggregation of information is the bottleneck, not labeling (\S\ref{sec:analysis}). We study the role of reasoning behavior and identify consistent error cases in reasoning models that limit their performance in this type of task, including over-reasoning by Gemini and DeepSeek models and task refusal by GPT models (\S\ref{sec:analysis}). 
We believe \ours is a usefully challenging evaluation of long-context reasoning abilities. 

\begin{table*}[htbp]
\centering
\small

\resizebox{\textwidth}{!}{\begin{tabular}{@{}l*{7}{c}@{}}
\toprule
& \multicolumn{3}{c}{\textbf{Real vs synthetic data}} & \multicolumn{2}{c}{\textbf{Aggregation capabilities}} & \multicolumn{2}{c}{\textbf{Measurability}} \\
\cmidrule(lr){2-4} \cmidrule(lr){5-6} \cmidrule(lr){7-8}
\textbf{Dataset} & \begin{tabular}[c]{@{}c@{}}Realistic\\inputs\end{tabular} & \begin{tabular}[c]{@{}c@{}}Realistic\\questions\end{tabular} & \begin{tabular}[c]{@{}c@{}}Relevant\\text matches\\distractors\end{tabular} & \begin{tabular}[c]{@{}c@{}}Multi-step\\reasoning\end{tabular} & Numeracy & \begin{tabular}[c]{@{}c@{}}Must\\use full\\input\end{tabular} & \begin{tabular}[c]{@{}c@{}}Easy to\\eval\end{tabular} \\
\midrule
MRCR & \cmark & \xmark & \cmark & \cmark & \cmark & \xmark & \cmark \\
BABILong & \xmark & \cmark & \xmark & \cmark & \xmark & \xmark & \cmark \\
RULER & \xmark & \xmark & \xmark & \smark & \smark & \smark & \cmark \\
\begin{tabular}[l]{@{}l@{}}HELMET Summ\end{tabular} & \cmark & \cmark & \cmark & \cmark & \xmark & \cmark & \xmark \\
\begin{tabular}[l]{@{}l@{}}HELMET LongQA\end{tabular} & \cmark & \cmark & \cmark & \cmark & \xmark & \smark & \cmark \\
\midrule
Oolong-synth & \xmark & \cmark & \cmark & \cmark & \cmark & \cmark & \cmark \\
Oolong-real & \cmark & \cmark & \cmark & \cmark & \cmark & \smark & \cmark \\
\bottomrule
\end{tabular}}

\vspace{0.5em}
\noindent Legend: \cmark~= yes, \xmark~= no, \smark~= in some cases
\caption{Comparing long-context reasoning benchmarks. \ours measures aggregation capabilities in  controllable, realistic, and simple-to-evaluate settings.}
\vspace{-0.5pt}
\label{tab:related_work_comp}
\end{table*}

\section{\ours-synth}
\label{sec:synth}

We aim to construct a synthetic information aggregation task that allows fine-grained control over the types of information and number of steps necessary to solve the task. To do this, we need building blocks: simple, short-context documents to build into a longer collection. We construct \ours-synth by constructing challenging corpus-level questions over existing in-context learning (ICL) datasets. This has conceptual similarities with database reasoning \citep{thorne-etal-2021-database}.


\subsection{Data}

We collect 10 text classification datasets with between 2 and 10 labels. We select for tasks that are possible for the authors to perform without difficulty (and validate that these tasks are similarly simple for models during our dataset filtering). \autoref{tab:icl_dataset_desc} describes the datasets in detail. We split the data into two validation tasks and eight test tasks. The validation tasks are selected to avoid underlying task overlap with test tasks (in the style of \citet{min2022metaicllearninglearncontext}), and so that there is both a 2-label and a multi-label (6-label) validation task.
For the sentiment and formality datasets with more than two labels, we reduce the label space to a single positive and a single negative label, to reduce the difficulty of the task. \\

\vspace{-15pt}
\paragraph{Data filtering}
In a normal in-context learning task, if a few examples are unusually hard or even mislabeled, this only slightly impacts on the overall score. Because we require aggregation across many ICL examples at once, one particularly difficult example can affect many answers downstream. However, many ICL datasets are known to contain mislabelings \citep{ying-thomas-2022-label, chong2022detectinglabelerrorsusing,annotation_error_detection}. We perform an additional screening step to remove mislabeled or unusually hard examples: we select two models that we do not expect to be substantially stronger than the models we are evaluating (GPT-4.1 nano and Llama 4 Maverick) and perform zero-shot ICL with a minimal instruction that provides the label space and type of task. We then exclude all examples that \textit{both} models get incorrect, for exclusion rates ranging from 0\% to 0.635\% by dataset. We report the percentage of examples screened out by dataset and provide examples of excluded and validated datapoints for each dataset in \autoref{app:icl_label_validation}. Many of the excluded instances are clearly mislabeled; the remainder generally represent challenging cases.

\subsection{Context window construction}
We determine the number of examples in the context window by using an estimate of the number of tokens required for each example, for the general task instructions, and for the specific question. Following \citet{yen2024helmet}, we compute these averages using the Llama 2 tokenizer, and use approximately 500K tokens of context for each estimate. We then estimate the number of examples to use by determining how many average-length examples would reach 95\% of the target context length.
We sample a distribution over the label classes so that the model cannot use any information about the true distribution over labels. 
Then, we sample examples to fill this distribution, and sample user IDs and dates.
Once a context window is constructed, we use it for 25 questions to allow for prompt caching, reducing the time to evaluate local models and the cost to evaluate on most APIs. At each context length, we sample two context windows per dataset, for a total of 50 questions per dataset or 400 test split questions per context length. We construct \ours-synth questions for every power of 2 from 1K to 4M. 

\paragraph{Questions}
We construct three types of \ours-synth questions, in order of increasing complexity. \textit{Counting} questions concern simple statistical properties of the label distribution, ranging from identifying the most frequent label to determining the number of examples with each label class. If the label distribution was provided, all tasks in this set should be trivial; these tasks can be seen implicitly as the task of labeling every example and reporting summary statistics. \textit{User information} questions require additional cross-reference with the \textit{user ID} field. \textit{Timeline} questions ask about changes in distribution before or after a certain date, between years, or between months across years. This is more challenging than user information questions because it requires reasoning about \textit{before} or \textit{after} a date, rather than matching a list of IDs (see \S\ref{app:app_oolong_synth} in Appendix for questions).

\paragraph{Instructions} For each dataset, we provide a brief (one-sentence) description of the task and label space, as well as the number of examples in-context. We provide instructions at the start and end of the input, as suggested by OpenAI's long context prompting guide.\footnote{\url{https://cookbook.openai.com/examples/gpt4-1\_prompting\_guide\#prompt-organization}} However, we provide the question only at the end, to enable prompt caching. 

\subsection{Evaluation}

\paragraph{Baseline} A random baseline is non-trivial for these tasks, because the output space for each question varies. We construct an algorithm for a random baseline with the following rules: (1) in cases where there is an $n$-way choice (e.g. choosing between $n$ labels or choosing a month where some criteria occurs), we select an answer from the set of valid answers at random; (2) if the question requires a numerical answer, we return $N/|L|$, where $N$ is the number of data points in context and $L$ is the labelset; (3) if the question requires a date or user ID, we sample at random from the list of dates/IDs in context. We compute the random baseline performance as the expected value of this procedure. 

\paragraph{Parsing answers} The task prompts specify an output format. We make a best-effort attempt to parse answers using this format; if this template is not present in the output, we take the candidate answer to be the last substring in the output that appears to match the desired answer type. 

\paragraph{Scoring}
For questions that require a label, date, user ID, or comparison (e.g. ``greater than'' or ``less than''), we score on exact match. For questions that require a numerical answer, we set the score to be $\texttt{score}(\hat y) = 0.75^{|y- \hat y|}$ to allow for partial credit.

\paragraph{An iterative system?}
Although \ours-synth allows for a high degree of control over the types of input, these inputs are composed of independent examples from the source dataset. In theory, an iterative system could complete these tasks by calling a model to label each individual example and then deterministically aggregating the resulting labels. However, such an iterative system could theoretically handle many existing evaluation tasks; for instance, Needle-in-a-Haystack tasks could be solved almost trivially by asking a model if each of the $N$ input sentences individually contains the needle.

We believe that the \ours-synth setting reflects the way many real-world users interact with current large language models. Real user queries tend to be underspecified and rarely use prompt engineering strategies \citep{sarkar2025conversationaluseraiinterventionstudy, xue2025userpromptingstrategieschatgpt}. When users do edit prompts, they are more likely to add additional context or more detailed instructions, rather than removing or subdividing tasks for the language model \citep{desmond2024exploringpromptengineeringpractices}. This use case is further exacerbated by recent systems supporting file uploads of arbitrary sizes for document analysis \citep{GoogleNotebookLM2025, AnthropicClaudeUpload2025}.
Asking questions on a large block of context represents a realistic use scenario, highlighting the utility of \ours-synth as a model capability evaluation bench. To further motivate the information aggregation problem, we next consider a setting in which inputs cannot be so trivially decomposed.

\section{\ours-real}
\label{sec:real}

We complement \ours-synth with questions derived from real conversational data. \ours-real is compiled from the transcripts of a Dungeons and Dragons (D\&D) role-playing game, where a group of players collaboratively build a story through in-character actions and success depends on rolling dice. Stories unfold over narratives (``campaigns'') that span dozens to hundreds of episodes, with each episode generally lasting 4-5 hours of play.  These transcripts involve several levels of conversation, from out-of-character chitchat to rules discussion to in-character actions and speech. Though lightly edited for readability, they reflect naturalistic speech instead of carefully planned written text. Because the conversation is unscripted and can involve tangents or side channels, conversational turns require variable amounts of prior context to resolve. In some instances, the same event (e.g. casting a spell) is discussed for many turns, or brought up again after a long interlude; in other cases, a prior event is ``retconned'' or revised post hoc.

In Dungeons and Dragons, there are limitations to how frequently characters can take certain types of actions
, and fans often discuss metadata of the campaigns (e.g., whether a certain person is particularly unlucky with their dice rolls or if a character uses a signature spell more or less as the campaign progresses). Because these shows are extremely long, wide-ranging, and improvised live, the creators do not provide this type of information. Instead, this information is often annotated meticulously, with multiple levels of verification, by dedicated fans of the work. 
We take this as an example of an information aggregation task in the wild that is \textit{not} simple to reframe as an iterative task.


\begin{figure*}[t!]
\centering
\begin{tikzpicture}
\pgfplotstableread[col sep=comma]{resources/plots/real_results.csv}\realdata
\pgfplotstabletranspose[colnames from=model]\realtable{\realdata}
\pgfplotstableread[col sep=comma]{resources/plots/synth_results.csv}\synthdata
\pgfplotstabletranspose[colnames from=model, columns={model,8192,16384,32768,65536,131072,262144,524288}]\synthtable{\synthdata}

\begin{groupplot}[
    group style={
        group size=2 by 1,
        horizontal sep=0.5cm,
    },
    legend style={
        at={(0.85,1.2)},
        anchor=south,
        legend columns=5,
        font=\scriptsize,
    },
    height=6cm,
]

\nextgroupplot[
    title={\ours-synth},
    xlabel={Context Length},
    ylabel={Score},
    grid=major,
    xmode=log,
    log basis x=2,
    xmin=6500, xmax=550000,
    ymin=0, ymax=1,
    xtick={8192,16384,32768,65536,131072,262144,524288},
    xticklabels={8K,16K,32K,64K,128K,256K,512K},
    width=8.5cm,
]

\addplot[
    color=cbYellow,
    mark=square,
    mark size=2pt,
    line width=1pt
] table[x=colnames, y={gemini-3-pro}] {\synthtable};
\addlegendentry{Gemini-3-Pro}

\addplot[
    color=cbBlue,
    mark=+,
    mark size=2pt,
    line width=1pt
] table[x=colnames, y={gpt-5}] {\synthtable};
\addlegendentry{GPT-5}

\addplot[
    color=cbOrange,
    mark=x,
    mark size=2pt,
    line width=1pt
] table[x=colnames, y={o3}] {\synthtable};
\addlegendentry{o3}

\addplot[
    color=cbGreen,
    mark=o,
    mark size=2pt,
    line width=1pt
] table[x=colnames, y={claude-sonnet-4-20250514}] {\synthtable};
\addlegendentry{Claude-Sonnet-4}

\addplot[
    color=cbPurple,
    mark=x,
    mark size=2pt,
    line width=1pt
] table[x=colnames, y={Llama3-3B}] {\synthtable};
\addlegendentry{Llama-3.2-3B}

\addplot[
    color=maroon,
    mark=o,
    mark size=2pt,
    line width=1pt
] table[x=colnames, y={Qwen3-4B-instruct}] {\synthtable};
\addlegendentry{Qwen3-4B-instruct}

\addplot[
    color=cbRed,
    mark=square,
    mark size=2pt,
    line width=1pt
] table[x=colnames, y={Olmo3.1-Think-32B}] {\synthtable};
\addlegendentry{Olmo-3.1-Think-32B}

\addplot[
    color=cbDarkYellow,
    mark=+,
    mark size=2pt,
    line width=1pt
] table[x=colnames, y={Gemma4-31B}] {\synthtable};
\addlegendentry{Gemma4-31B}

\addplot[
    color=cbGray,
     mark=o,
     mark size=2pt,
     line width=1pt
] table[x=colnames, y={deepseek-r1-0528}] {\synthtable};
\addlegendentry{Deepseek-R1-0528}

\addplot[
    color=black,
    mark=.,
    dashed,
    mark size=2pt,
    line width=1pt
] table[x=colnames, y={random-baseline}] {\synthtable};
 \addlegendentry{random baseline}

\nextgroupplot[
    title={\ours-real},
    legend entries={},
    xlabel={Context Length},
    xmode=log,
    log basis x=2,
    xmin=48000, xmax=600000,
    ymin=0, ymax=1,
    xtick={32768,65536,131072,262144,524288},
    xticklabels={32K,64K,128K,256K,512K},
    yticklabel pos=right,
    grid=major,
    width=6cm,
    ]

\addplot[
    color=cbYellow,
    mark=square,
    mark size=2pt,
    line width=1pt
] table[x=colnames, y={gemini-3-pro}] {\realtable};

\addplot[
    color=cbBlue,
    mark=+,
    mark size=2pt,
    line width=1pt
] table[x=colnames, y={gpt-5}] {\realtable};

\addplot[
    color=cbOrange,
    mark=x,
    mark size=2pt,
    line width=1pt
] table[x=colnames, y={o3}] {\realtable};

\addplot[
    color=cbGreen,
    mark=o,
    mark size=2pt,
    line width=1pt
] table[x=colnames, y={claude-sonnet-4-20250514}] {\realtable};

\addplot[
    color=cbPurple,
    mark=x,
    mark size=2pt,
    line width=1pt
] table[x=colnames, y={Llama3-3B}] {\realtable};
%

\addplot[
    color=maroon,
    mark=o,
    mark size=2pt,
    line width=1pt
] table[x=colnames, y={Qwen3-4B-instruct}] {\realtable};
%

\addplot[
    color=cbRed,
    mark=square,
    mark size=2pt,
    line width=1pt
] table[x=colnames, y={Olmo3.1-Think-32B}]{\realtable};
%

\addplot[
    color=cbDarkYellow,
    mark=+,
    mark size=2pt,
    line width=1pt
] table[x=colnames, y={Gemma4-31B}] {\realtable};

\addplot[
    color=cbGray,
    mark=o,
    mark size=2pt,
    line width=1pt
] table[x=colnames, y={deepseek-r1-0528}] {\realtable};

\end{groupplot}

\end{tikzpicture}
\caption{Scores by context window length for \ours-synth and \ours-real. Here we show only the best model for each model family; see Appendix~\ref{app:results} for full results.}
\label{fig:results_by_length}
\vspace{-4pt}
\end{figure*}


For \ours-real, we consider two campaigns from the series Critical Role and the data compiled by CritRoleStats,\footnote{\url{https://www.critrolestats.com/}} a fan project that labeled by hand per-episode statistics for the first several campaigns of Critical Role. We used episode transcripts from the Critical Role Dungeons and Dragons Dataset (CRD3) \citep{rameshkumar-bailey-2020-storytelling}. 
We use CritRoleStats game statistics about dice rolls and spells cast in each episode and design a set of questions that cover a variety of information aggregation tasks. We include questions that require processing of single- or multiple-episode transcripts. For multi-episode questions, we concatenate transcripts and use delimiters to highlight the start and end of each transcript. See \autoref{tab:oolong_real_single}, \autoref{tab:oolong_real_multi_rolls}, and \autoref{tab:oolong_real_multi_spells} of the Appendix for a full list of \ours-real questions. We include varying context windows to evaluate the model's ability to use long-context reasoning and aggregation capabilities. Using a single episode transcript as a context unit, we include context windows ranging from one to 24 episode transcripts (55K to 1.3M tokens). We follow the evaluation setup from \ours-synth.

\ap{add a line about how question design ensures multi-piece evidence}


\ap{discuss how oolong-real connects to similar use cases in finance and medical domains.}

\section{Results and Analysis}
\label{sec:analysis}

On \ours, we benchmark a strong suite of frontier models to study their information aggregation capabilities over long context. 
We include a mix of models of varying sizes, levels of reasoning, and long-context capabilities.
In \autoref{fig:results_by_length}, we visualize the model performance for context windows up to 512K tokens. As expected, we see a significant drop in performance at higher context windows. The two splits are of similar difficulty at the same context length, although direct comparison is challenging because real data do not necessarily align with fixed context window buckets.\footnote{While performance on \ours-real starts lower, this is because the shortest inputs in this dataset are a single episode, with an average length of 55K tokens; models perform similarly on \ours-synth around the same context length.} A breakdown of results with an aggregate rating over the two datasets is available in the Appendix (\autoref{tab:leaderboard}), as well as an analysis by question and answer type (\autoref{app:question_type_analysis}).

\subsection{Does reasoning effort correlate to performance?}
\label{ssec:analysis_reasoning}

\ours is a reasoning-intensive task, so we hypothesize that increased reasoning may correlate with better scores. We measure reasoning effort in three different ways. 

\begin{figure}
\begin{subfigure}{0.45\textwidth}
    \centering
    \begin{tikzpicture}[scale=0.7]
\begin{axis}[
    title={\ours-synth},
    xlabel={Context Length},
    ylabel={Score},
    grid=major,
    width=9cm,
    height=6cm,
    xmode=log,
    log basis x=2,
    xmin=6500, xmax=512000,
    ymin=0, ymax=1,
    xtick={8192,16384,32768,65536,131072,262144,524288},
    xticklabels={8K,16K,32K,64K,128K,256K,512K},
]

\pgfplotstableread[col sep=comma]{resources/plots/synth_results.csv}\datatable
\pgfplotstabletranspose[colnames from=model, columns={model,8192,16384,32768,65536,131072,262144,524288}]\transposedtable{\datatable}

 \addplot[
     color=cbBlue,
     mark=*,
     mark size=2pt,
     line width=1pt
 ] table[x=colnames, y={gpt-5-nano}] {\transposedtable};
 \addlegendentry{gpt-5-nano [default]} 

\addplot[
     color=cbOrange,
     mark=x,
     mark size=2pt,
     line width=1pt
 ] table[x=colnames, y={gpt-5-nano-high}] {\transposedtable};
 \addlegendentry{gpt-5-nano [high]}

\addplot[
     color=cbGreen,
     mark=o,
     mark size=2pt,
     line width=1pt
 ] table[x=colnames, y={gpt-5-nano-low}] {\transposedtable};
 \addlegendentry{gpt-5-nano [low]}
 
\end{axis}
\end{tikzpicture}
    \caption{GPT-5-nano reasoning effort}
    \label{fig:ablation_reasoning_levels_synth}
\end{subfigure}
\hfill
\begin{subfigure}{0.45\textwidth}
\centering
    \includegraphics[width=0.9\linewidth]{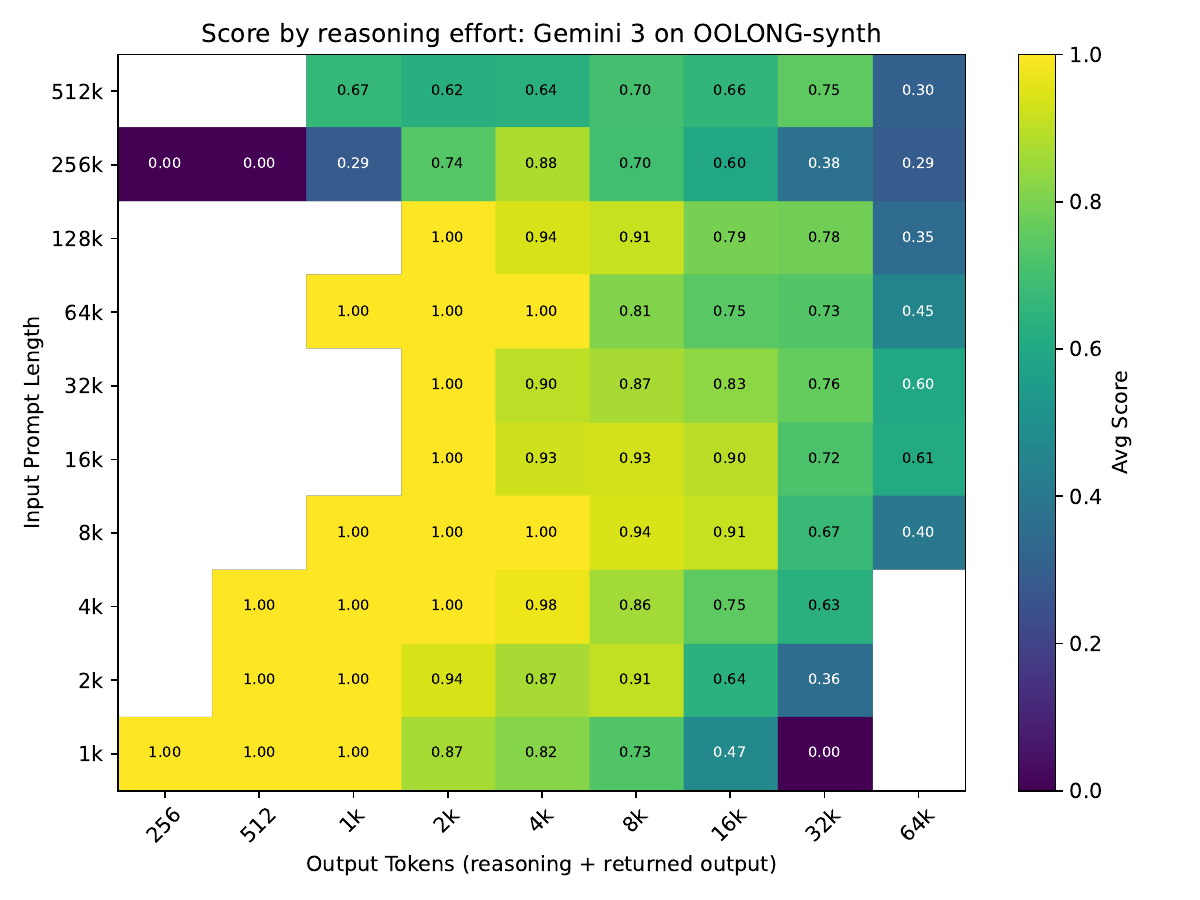}
    \caption{Gemini-3-Pro reasoning token count}
    \label{fig:heatmap}
\end{subfigure}
\caption{Does reasoning effort correlate with task performance?}
\end{figure}
\paragraph{Qwen and Olmo: think versus instruct models.} In these families, we have pairs of models that are focused on reasoning and conventional instruction following. In the Olmo family, the reasoning (``think'') models greatly exceed the performance of their equivalent instruction-tuned model on \ours-synth. However, for Qwen-3-4B, the reasoning model shows near-zero performance on \ours-synth because of the same over-reasoning phenomenon described for Deepseek-R1: although it is allotted up to 256K output tokens, the model almost always uses the full output for reasoning without returning an answer. 

\paragraph{GPT-5-nano: reasoning effort parameter.} The OpenAI API allows for specification of a reasoning level: ``low,'' ``medium,'' or ``high.'' We compare ``high'' and ``low'' reasoning effort for GPT-5-nano. \autoref{fig:ablation_reasoning_levels_synth} shows the results for \ours-synth. Although \ours is a reasoning-intensive task, specifying a higher reasoning effort is only useful for short contexts; after 64k, there is little discernible difference between reasoning levels, with the ``high'' reasoning setting even slightly underperforming ``low'' reasoning at 256k on \ours-synth. 
We hypothesize that at context lengths where there is sufficient remaining room in the context window to enumerate labels for each example in-context, adding more reasoning effort may encourage the model to take this strategy. 
However, since the default routing for this model performs reasonably well at all lengths and especially for longer inputs, we do not explicitly specify a reasoning level for the remainder of the runs. We report results on \ours-real in \autoref{fig:ablation_reasoning_levels_real} in the Appendix.


\paragraph{Gemini 3 Pro: reasoning token count.} The Gemini API returns the number of reasoning tokens used (but not the content of the reasoning trace). We visualize trends in reasoning tokens  used in \autoref{fig:heatmap}. Longer inputs generally lead to longer reasoning traces, as expected; however, when the input length is held constant, reasoning trace length correlates \textit{negatively} with correctness. We hypothesize this could partially be because harder problems result in both longer traces and lower probability of answering correctly. 




\subsection{Over-reasoning}
We observe a persistent failure case related to \textit{over-reasoning} on several models: the model exceeds its maximum reasoning token or total token budget in the course of a very long reasoning trace, and fails to answer the question at all.\footnote{We evaluate all models on their allowed maximum reasoning trace or output token budget.}  This occurs particularly on \ours-synth, which is extremely information-dense--- all or nearly all of the lines are necessary to solve the questions. In our experiments, over-reasoning is the culprit when a model performs well on \ours-real and poorly on \ours-synth. 

\paragraph{Deepseek R1} This is a strong reasoning model and outperforms GPT-5 nano and o4-mini on \ours-real. However, on \ours-synth, it achieves performance below the random baseline. 
In manual inspection of the traces, we observe pathologies in the reasoning traces for \ours-synth that are not present in traces for \ours-real and appear only occasionally in traces from other models. We identify these behaviors through discussion between the authors, then prompt GPT-5.6-Luna to label 2,400 randomly selected traces from Deepseek-R1 with a yes/no classification about whether an identified behavior is present in this trace.\footnote{For this labeling task, we also verify with a human annotation of a small subset, and see high agreement (85-96\%) across all questions.} Many traces (56.3\%) do not provide an answer at all. We hypothesize that, because \ours-synth is such an information-dense task, the model's apparent strategy of labeling each example before deciding which are relevant results in running out of context tokens. 75.7\% of traces end in an incomplete sentence.\footnote{Note that in some cases, the model provides a candidate answer, then begins to double-check the answer and then exceeds the maximum output length; we evaluate the last candidate answer we are able to parse.} In 26.6\% of cases, the model debates whether the task is impossible or intractable given the length of the input. 
\autoref{app:deepseek_example_traces} shows example traces for Deepseek-R1 from both splits. 

\paragraph{Gemini models} Gemini 2.5 Pro is one of the strongest models on \ours-real, but exhibits performance dropoff at longer contexts on \ours-synth; Gemini 3 experiences the same trend less sharply, falling from the best model by a margin to comparable with Claude Sonnet by 256K on \ours-synth. We observe that both models, and especially Gemini 2.5 Pro, increasingly exceeds its maximum output length during reasoning for long \ours-synth inputs. 
Unlike the other APIs used in benchmarking, the Gemini API does not return any tokens if the max token count is exceeded during reasoning; this results in an automatic score of 0 for any cases of overlong reasoning.
At 256K context length, Gemini 2.5 Pro fails to return an answer because of the max reasoning tokens constraint in 64.0\% of \ours-synth outputs, despite these inputs requiring only one quarter of the model's input context window. At 256K, Gemini 3 Pro fails to return 45\% of \ours-synth outputs. Gemini 3 Pro outperforms the older model by both answering more queries and answering a higher percentage of them correctly.
\footnote{Note the slight increase in Gemini 3 Pro performance from 256K to 512K: while 45\% of inputs at 256K hit the max token limit during reasoning, only 11.8\% of inputs at 512K hit this limit. It is not clear what causes this shift.} Over-reasoning is not the only reason for declining performance with longer context; however, it bounds performance for both Gemini models. 
We do not observe this behavior on \ours-real, where Gemini maintains its strong performance at longer context lengths. 


\subsection{Refusal}
Another error case we observe is when models \textit{refuse} to answer a question at all. In these responses, models describe the task as impossible or intractable.
Certainly some level of caution is appropriate when answering types of questions that models struggle on. However, we compare the performance of GPT 5 and GPT 5.2 and observe that surprisingly, despite very strong performance on OpenAI's MRCR long context benchmark \cite{openai-2025-mrcr}, GPT 5.2 \textit{underperforms} GPT 5 substantially on \ours-synth. This is because GPT 5.2 is much more likely to refuse to perform the task; at 256K, GPT-5 attempts 83\% of queries and answers 44.5\% of these correctly, while GPT-5.2 attempts only 32\% with an accuracy of 52.8\%. It seems likely that GPT-5.2's performance would be improved, perhaps substantially, by attempting a higher percentage of questions.

\subsection{Aggregation without classification} 

To isolate the effects of aggregation and classification, we consider simpler settings for both dataset splits. In both settings, we see that aggregation is the bottleneck: scores are only marginally improved by removing the classification task.

\paragraph{Providing labels} 
\begin{wrapfigure}{r}{0.5\textwidth}
    \centering
    \begin{tikzpicture}[scale=0.9]
\centering
\begin{axis}[
    xlabel={Context Length},
    ylabel={Score},
    grid=major,
    width=7.5cm,
    height=6cm,
    xmode=log,
    log basis x=2,
    xmin=900, xmax=80000,
    ymin=0, ymax=1,
    xtick={1024,2048,4096,8192,16384,32768,65536},
    xticklabels={1K,2K,4K,8K,16K,32K,64K},
    legend style={
        font=\scriptsize,
        at={(0.5,-0.25)},
        anchor=north,
        legend columns=2,
        /tikz/column sep=1ex,
    },
]

\pgfplotstableread[col sep=comma]{resources/plots/synth_results.csv}\datatable
\pgfplotstabletranspose[colnames from=model, columns={model,1024,2048,4096,8192,16384,32768,65536}]\transposedtable{\datatable}

\addplot[
    color=cbOrange,
    mark=*,
    mark size=2pt,
    line width=1pt
] table[x=colnames, y={gpt-5-labels}] {\transposedtable};
\addlegendentry{GPT-5 +labels}

\addplot[
     color=cbOrange,
     mark=o,
     mark size=2pt,
     line width=1pt
 ] table[x=colnames, y={gpt-5}] {\transposedtable};
 \addlegendentry{GPT-5}

\addplot[
    color=cbBlue,
    mark=*,
    mark size=2pt,
    line width=1pt
] table[x=colnames, y={gpt-5-nano-labels}] {\transposedtable};
\addlegendentry{GPT-5-nano +labels}

\addplot[
    color=cbBlue,
    mark=o,
    mark size=2pt,
    line width=1pt
] table[x=colnames, y={gpt-5-nano}] {\transposedtable};
\addlegendentry{GPT-5-nano}

 \addplot[
      color=cbYellow,
      mark=*,
      mark size=2pt,
      line width=1pt
  ] table[x=colnames, y={gemini-2.5-flash-labels}] {\transposedtable};
  \addlegendentry{Gemini 2.5 flash +labels}

  \addplot[
      color=cbYellow,
      mark=o,
      mark size=2pt,
      line width=1pt
  ] table[x=colnames, y={gemini-2.5-pro}] {\transposedtable};
  \addlegendentry{Gemini 2.5 flash} 

  \addplot[
      color=cbGreen,
      mark=*,
      mark size=2pt,
      line width=1pt
  ] table[x=colnames, y={claude-sonnet-labels}] {\transposedtable};
  \addlegendentry{Sonnet 4 +labels}

  \addplot[
      color=cbGreen,
      mark=o,
      mark size=2pt,
      line width=1pt
  ] table[x=colnames, y={claude-sonnet-4-20250514}] {\transposedtable};
  \addlegendentry{Sonnet 4}
  
\end{axis}
\end{tikzpicture}
\label{fig:synth_ablation_labels}
    \caption{Comparison on \ours-synth: we provide the gold labels in the input. This leads to a consistent but small improvement.}
    \label{fig:ablation_labels}
\end{wrapfigure}
For \ours-synth, we construct an easier version of the task by providing the label for each ICL example in-context. 
This reduces the task for most questions to identifying the relevant instances and summing the occurrence of each label type. \autoref{fig:ablation_labels} shows the behavior of four models with and without labels provided in-context. As expected, adding labels improves accuracy; however, this improvement ranges from 10.9 points to only 0.79 points.  
We do not see consistently higher gains from providing labels in longer inputs, which suggests that the lower performance at longer context lengths is not primarily due to an accumulation of mislabeling errors.
 Additionally, the improvement is not larger for GPT-5-nano than it is for the more powerful GPT-5, suggesting that the performance difference observed between these models is not due to differing performance on the classification task. This is by design; our classification task validation was designed to eliminate overly challenging or misleading examples. 

\paragraph{Providing domain knowledge} We believe that most \ours-real tasks do not require knowledge specific to Dungeons \& Dragons. For instance, tasks that require counting occurrences of a spell or action do not require knowledge of the goals of that action, only its presence in the text. Some questions, however, do benefit from domain knowledge; for instance, the question ``how many cantrips were cast?'' requires knowing which spells are considered cantrips, and the question ``which roll type is most common?'' is solvable without external knowledge but may be easier with the context of all the types of rolls and where each is generally used. While models generally have this knowledge because of the proliferation of text about this game on the internet, we experiment with providing additional context for each of these questions: a mapping from spell name to spell level or a list of roll types with brief descriptions of each. In this setting, we rerun the single-episode examples for GPT-5 and Claude Sonnet 4 and observe that providing this additional context helps slightly but not substantially: raising GPT-5’s score by 3.51 points to 62.25 and raising Claude Sonnet 4’s score by 2.57 points to 53.15. 

\subsection{Iterative retrieval baseline for \ours-real}
While \ours-real remains an information-dense task, with dozens of individual details needed to answer most questions, there are segments of the transcripts that are not necessary to generate the answer. 
Could \ours-real be tackled with retrieval? 

We use \texttt{smolagent} \citep{smolagents} to instantiate a simple iterative RAG baseline acting for up to 5 steps. This system is equipped with tool calls for retrieval (at an utterance level, using a BM25 retriever) and code execution. Each utterance is annotated with the transcript number and the utterance number before being indexed. We evaluated this system on the single transcript and two transcript settings (the 55K and 110K context settings in \autoref{fig:results_by_length}) with GPT-5 as the backbone model. This system substantially \textit{underperforms} directly using GPT-5 (with one transcript: -16.15 points; with two transcripts: -17.67 points). 
This shows that \ours cannot be easily solved by a simple agentic retrieval system. We expect future work to study ways to improve retrieval tool use and explore dynamic chunking algorithms depending on the question. The interleaved and complex nature of human dialogue presents unique challenges to both agentic and long-context systems alike (see \autoref{app:dnd_data_rocks} for further discussion). However, retrieval performance, while below long-context performance, remains non-trivial; because of this, we speculate that methods that extend context by condensing sections of the context \citep{smith2025openhands} or evicting subsections of the context \citep{li2025surveylargelanguagemodel} could be effective for this task. 

\section{Related Work}
\label{sec:related_work}
We are far from the first to recognize the importance of aggregation tasks. 
Many benchmarks focus on reasoning tasks in settings where longer context is used to measure robustness to distractors, rather than to introduce increased numbers of reasoning steps. This is an important problem but orthogonal to our focus. For instance, BABILong \citep{kuratov2024babilongtestinglimitsllms} tests long-context reasoning by placing BABI \citep{babi} reasoning tasks within a long context and evaluating performance as the amount of distractor text increases. \citet{ling2025longreasonsyntheticlongcontextreasoning}, a contemporaneous work to ours, creates longer-context reasoning problems by synthetically elaborating on the documents necessary for short-context reasoning tasks and then interleaving these with irrelevant context. 

Other tasks require use of the full context in heavily artificial settings. RULER \citep{hsieh-2024-ruler} includes synthetic retrieval tasks at varying context lengths. It includes multi-hop tracing and aggregation tasks such as identifying the most common words in a long input of random nouns. LongMemEval \citep{wu-2025-longmemeval} and multi-round coreference resolution tasks (MRCR) evaluate long-context capabilities with conversational data. 
MRCR was first introduced in Gemini \citep{vodrahalli-2024-michelangelo} and was further extended in OpenAI's MRCR \citep{openai-2025-mrcr}. Unlike the standard NIAH task, the needles and the distractors in MRCR are sampled from the same distributions. OpenAI's MRCR includes variants of 2, 4, and 8 needles, and the task is to retrieve the \emph{i}th instance of one of the needles. While this uses naturalistic data (synthetic conversations generated using GPT-4o), it is a task focused on measuring basic counting ability rather than reasoning. Conversely, GSM-infinite  \citep{zhou2025gsminfinitellmsbehaveinfinitely} stress tests long-context reasoning through adaptable frameworks for varying task difficulty and length; our work is complementary, as their focus is primarily on mathematical reasoning through synthetically generated problems, while we focus on a more noisy, naturalistic text setting with primarily non-mathematical reasoning. Similarly, MathHay \citep{wang2024mathhayautomatedbenchmarklongcontext} focuses on queries across document collections that require retrieving multiple numbers from across context and performing multi-step mathematical reasoning with them.

Multi-hop or long context question answering tasks can also require collecting information from multiple parts of the context, and represent a different type of long context reasoning task: one much sparser and less numeracy-focused than our aggregation task. 
HELMET \citep{yen2024helmet} expands the tasks in RULER to include tasks related to the downstream use of LMs  
including a set of long-context QA tasks from \citet{zhang2024inftybenchextendinglongcontext}. Other long-context QA tasks explicitly focus on capturing long dependency length \citep{li-etal-2024-loogle, bai2025longbenchv2deeperunderstanding}, on building semantic equivalents to needle-in-a-haystack retrieval \citep{modarressi2025nolimalongcontextevaluationliteral}, or specific downstream domains \citep{he2025looglev2llmsready}
TLDM \citep{hamilton-2025-too-long} includes two related tasks about character location and time passed in a narrative setting in the domain of public-domain literature. 
MoNaCo \citep{wolfson-etal-2025-monaco} considers challenging information seeking queries that require retrieval and aggregation; their emphasis is on retrieval of the necessary documents for intermediate reasoning steps in an agentic setting.

Other works conceptually aligned with ours include ZeroScrolls and NeedleBench. 
ZeroSCROLLS \citep{shaham2023zeroscrollszeroshotbenchmarklong} introduces a task of identifying the percentage of reviews on Amazon for a product that are positive, which is the closest conceptual ancestor of our work; \ours-synth encompasses this type of task along with other aggregation tasks. NeedleBench \citep{li2025needlebenchevaluatingllmretrieval} introduces a reasoning task called the Ancestral Trace Challenge, which, given a number of relations between individuals in context, asks the model to identify the most distance relative of a specific individual. The Ancestral Trace Challenge requires tree-like branching of reasoning; \ours focuses on classify-then-aggregate style tasks and emphasizes more realistic data. 

A line of work has also framed LLM aggregation as database-like operations over the input text. \citet{thorne-etal-2021-database} call this problem database reasoning and study architectures well-suited to this task for short contexts. HoloBench \citep{maekawa2025holisticreasoninglongcontextlms} directly synthesizes examples from text-to-SQL datasets.
 And concurrently to our work, SciTrek \citep{li2026scitrekevaluatingimprovinglongcontext} frames SQL-like queries over scientific article metadata as a long context evaluation. 

\autoref{app:related_work} discusses additional related work on information aggregation and on Dungeons \& Dragons data.

\section{Conclusion}
We introduce \ours, a challenging long-context information aggregation benchmark in two parts. 
Frontier models' poor performance on \ours suggests more work is necessary in developing both reasoning abilities and more general long context abilities. 
In longer context regimes, many models perform roughly equivalently on \ours-synth and \ours-real at the same context length, despite the much higher information density in \ours-synth. This suggests that better length generalization, regardless of reasoning ability, may also improve performance on this type of challenging task. 
However, some models behave differently in extremely information-dense regimes, highlighting a direction for future work. 
While prior work on reasoning chains has focused on over-reasoning for adversarial \citep{kumar2025overthinkslowdownattacksreasoning} or overly simple problems \citep{chiang2024overreasoningredundantcalculationlarge, sui2025stopoverthinkingsurveyefficient}, \ours indicates that even strong models may struggle to plan the reasoning quantity for information-dense inputs, where it may be desirable to accept a less optimal or more error-prone reasoning strategy (e.g. double-checking less frequently) in order to avoid running into the maximum reasoning tokens. Future models that are aware of the maximum reasoning token budget may be able to more carefully allocate reasoning effort. 
Overall, our results suggest that there is still a long way to go in designing robust long-context aggregation capabilities for LLMs.

\section*{Reproducibility Statement}
We release code for dataset construction and evaluation, in addition to the final dataset version. API-based models typically exhibit some small variability across API calls, and versions of each model may be phased out over time by the providers, so exact replication of outputs may be infeasible; wherever possible, we release the model outputs for each model run reported in the paper. 

\section*{Ethics Statement}
This work is intended to further the evaluation of long-context LLMs. We do not foresee any particular additional risks introduced by our evaluation, although information aggregation, like nearly any LM capability, could be useful for both beneficial and harmful actors. 

CritRoleStats, whose data we use as part of constructing gold labels for the \ours-real split, explicitly allows the use of their data for statistical analyses in their FAQ, so long as they are credited.\footnote{\url{https://www.critrolestats.com/faqs}} We have also notified them of our use of their data.

In the course of this work, we used LM assistance in the writing of some data preprocessing, analysis, and visualization scripts.

\section{Acknowledgements}

We are grateful to the contributors and maintainers of CritRoleStat for their excellent resource, and to the broader Critical Role community for their commitment and engagement. Like prior work that uses Dungeons \& Dragons as a testbed for capabilities, this work would not be possible without the resources built by fan communities. 
We would also like to thank Clara Na for helpful feedback on drafts of this work. 

 AB was supported by a grant from the National Science Foundation Graduate Research Fellowship Program under Grant No. DGE2140739. Any opinions, findings, and conclusions or recommendations expressed in this material are those of the author(s) and do not necessarily reflect the views of the sponsors.

\newpage

\bibliography{references}
\bibliographystyle{colm2026_conference}

\newpage 
\appendix

\section{\ours – Leaderboard}
Here we provide full results on all models we ran.

In \autoref{tab:leaderboard}, we report the results for both \ours-synth and \ours-real numerically. Following prior long-context benchmarks, we measure model performance at increasing context window sizes and report a task average. For \ours-synth, we use the standard context window sizes of up to 128K tokens. For \ours-real, we report scores at increasing number of episode transcripts (up to 3 episodes). GPT-5 performs the best in both subsets, followed by o3, GPT-5-mini and Claude-Sonnet-4. Gemini-2.5-Pro is initially the strongest model, but falls to below-random performance by 256K, largely because of responses exceeding the maximum token count; see the analysis for further discussion. Notably, at 128K context window size nearly all our evaluated models obtain a score under 50. DeepSeek R1 does well on \ours-real, but struggles to perform well on \ours-synth (see \S\ref{ssec:analysis_reasoning}. Llama-4-Maverick fails in both subsets.

\begin{table*}[t]
\centering
\resizebox{0.85\textwidth}{!}{
\begin{tabular}{lr|rrrrr !{\vrule width 1.5pt} r |rrr}
\multicolumn{1}{c}{} & \multicolumn{6}{c}{\textbf{\ours-synth}} & \multicolumn{4}{c}{\textbf{\ours-real}} \\
\cmidrule(lr){2-7} \cmidrule(lr){8-11}
Model &  Avg. & 8K & 16K & 32K & 64K & 128K & Avg. & 55K & 118K & 175K \\
\midrule
Gemini-3-Pro & \textbf{78.69} & \textbf{90.94} & \textbf{87.61} & \textbf{81.94} & \textbf{70.55} & \textbf{62.42} & \textbf{57.93} & \textbf{64.71} & \textbf{55.01} & \textbf{54.07} \\
GPT-5 & 70.75 & 85.56 & 84.45 & 76.12 & 61.24 & 46.36 & 47.00 & 58.74 & 45.72 & 36.53 \\
Gemini-2.5-Pro & 55.29 & 88.13 & 69.84 & 56.83 & 36.56 & 25.06 & 52.16 & 60.04 & 49.36 & 47.07 \\
o3 & 62.37 & 86.80 & 79.52 & 63.23 & 44.86 & 37.45 & 36.71 & 50.57 & 33.57 & 25.99 \\
GPT-5-mini & 63.68 & 85.13 & 77.65 & 64.64 & 50.14 & 40.85 & 34.55 & 49.86 & 29.90 & 23.89 \\
Claude-Sonnet-4 & 58.18 & 74.43 & 62.75 & 55.67 & 50.04 & 48.02 & 36.75 & 50.58 & 32.98 & 26.70 \\
o4-mini & 56.74 & 83.07 & 65.10 & 51.86 & 44.15 & 39.53 & 27.13 & 41.69 & 21.77 & 17.93 \\
Gemma-4-31B & 53.38$^\dagger$ & 79.29 & 69.80 & 64.32 & 53.47 & -- & 30.23$^\dagger$ & 54.97 & 35.72 & -- \\
GPT-5-nano & 50.73 & 70.96 & 54.53 & 47.81 & 41.02 & 39.31 & 31.05 & 43.09 & 26.82 & 23.23 \\
GPT-5.2 & 44.29 & 61.80 & 53.95 & 47.32 & 35.31 & 23.10 & 34.40 & 45.67 & 32.15 & 25.38 \\
Qwen3-4B-Instruct & 40.67 & 55.81 & 41.33 & 41.35 & 34.36 & 30.49 & 22.88 & 39.58 & 18.29 & 10.78 \\
Olmo-3.1-Think-32B & 35.29$^\dagger$ & 59.01 & 43.81 & 39.13 & 34.48 & -- & 10.43$^\dagger$ & 31.28 & -- & -- \\
Deepseek-R1 & 13.11 & 13.94 & 13.65 & 12.91 & 13.20 & 11.87 & 32.00 & 47.85 & 27.35 & 20.81 \\
Olmo-3-Think-7B & 31.42$^\dagger$ & 46.44 & 39.22 & 38.14 & 33.32 & -- & 7.47$^\dagger$ & 22.42 & -- & -- \\
Olmo-3.1-Instruct-32B & 26.67$^\dagger$ & 42.31 & 38.35 & 31.64 & 21.06 & -- & 10.35$^\dagger$ & 31.04 & -- & -- \\
Olmo-3-Instruct-7B & 21.10$^\dagger$ & 29.88 & 28.55 & 25.54 & 21.54 & -- & 7.93$^\dagger$ & 23.80 & -- & -- \\
Llama-3-3B & 25.60 & 29.31 & 26.37 & 24.86 & 25.09 & 22.39 & 0.00 & 0.00 & 0.00 & 0.00 \\
Llama-3-8B & 23.71 & 27.54 & 24.66 & 24.74 & 24.29 & 17.31 & 0.41 & 1.15 & 0.07 & 0.00 \\
Qwen3-4B-Thinking & 0.00 & 0.00 & 0.00 & 0.00 & 0.00 & 0.00 & 20.18 & 37.65 & 15.75 & 7.15 \\
Llama-4-Maverick & 16.37 & 15.00 & 16.29 & 15.42 & 16.35 & 18.80 & 2.07 & 2.48 & 2.11 & 1.62 \\
\bottomrule
\end{tabular}}
\caption{\ours results on a number of strong models. Nearly all models we test support at least 200K context; thus, we report \ours scores as an average over scores on 8K-175K inputs. \ours-synth is a more information-dense task than \ours-real; the ranking of some models shifts slightly between the two settings. Models are sorted by the average score between the two splits. --~marks a context length beyond the maximum for that model; such lengths count as 0 in the average, which we denote with $^\dagger$.}
\label{tab:leaderboard}
\end{table*}
Figure~\ref{fig:closed_models} and Figure~\ref{fig:open_models} are augmented versions of Figure~\ref{fig:results_by_length} with all closed-weights and open-weights models (respectively) shown.

\begin{figure*}[]
\centering
\begin{tikzpicture}
\pgfplotstableread[col sep=comma]{resources/plots/real_results.csv}\realdata
\pgfplotstabletranspose[colnames from=model]\realtable{\realdata}
\pgfplotstableread[col sep=comma]{resources/plots/synth_results.csv}\synthdata
\pgfplotstabletranspose[colnames from=model, columns={model,8192,16384,32768,65536,131072,262144,524288}]\synthtable{\synthdata}

\begin{groupplot}[
    group style={
        group size=2 by 1,
        horizontal sep=0.5cm,
    },
    legend style={
        at={(0.85,1.2)},
        anchor=south,
        legend columns=5,
        font=\scriptsize,
    },
    height=6cm,
]

\nextgroupplot[
    title={\ours-synth},
    xlabel={Context Length},
    ylabel={Score},
    grid=major,
    xmode=log,
    log basis x=2,
    xmin=6500, xmax=550000,
    ymin=0, ymax=1,
    xtick={8192,16384,32768,65536,131072,262144,524288},
    xticklabels={8K,16K,32K,64K,128K,256K,512K},
    width=8.5cm,
]

\addplot[
    color=cbYellow,
    mark=o,
    mark size=2pt,
    line width=1pt
] table[x=colnames, y={gemini-2.5-pro}] {\synthtable};
\addlegendentry{gemini-2.5-pro}

\addplot[
    color=cbYellow,
    mark=square,
    mark size=2pt,
    line width=1pt
] table[x=colnames, y={gemini-3-pro}] {\synthtable};
\addlegendentry{gemini-3-pro}

\addplot[color=cbBlue, mark=o, mark size=2pt, line width=1pt]
  table[
    y={gpt-5-nano},
  ] {\synthtable};
\addlegendentry{gpt-5-nano}

\addplot[
    color=cbBlue,
    mark=x,
    mark size=2pt,
    line width=1pt
] table[x=colnames, y={gpt-5-mini}] {\synthtable};
\addlegendentry{gpt-5-mini}

\addplot[
    color=cbBlue,
    mark=+,
    mark size=2pt,
    line width=1pt
] table[x=colnames, y={gpt-5}] {\synthtable};
\addlegendentry{gpt-5}

\addplot[
    color=cbBlue,
    mark=square,
    mark size=2pt,
    line width=1pt
] table[x=colnames, y={gpt-5.2}] {\synthtable};
\addlegendentry{gpt-5.2}

\addplot[
    color=cbOrange,
    mark=o,
    mark size=2pt,
    line width=1pt
] table[x=colnames, y={o4-mini}] {\synthtable};
\addlegendentry{o4-mini}

\addplot[
    color=cbOrange,
    mark=x,
    mark size=2pt,
    line width=1pt
] table[x=colnames, y={o3}] {\synthtable};
\addlegendentry{o3}

\addplot[
    color=cbGreen,
    mark=o,
    mark size=2pt,
    line width=1pt
] table[x=colnames, y={claude-sonnet-4-20250514}] {\synthtable};
\addlegendentry{claude-sonnet-4}

\addplot[
    color=black,
    mark=.,
    dashed,
    mark size=2pt,
    line width=1pt
] table[x=colnames, y={random-baseline}] {\synthtable};
 \addlegendentry{random baseline}

\nextgroupplot[
    title={\ours-real},
    legend entries={},
    xlabel={Context Length},
    xmode=log,
    log basis x=2,
    xmin=48000, xmax=600000,
    ymin=0, ymax=1,
    xtick={32768,65536,131072,262144,524288},
    xticklabels={32K,64K,128K,256K,512K},
    yticklabel pos=right,
    grid=major,
    width=6cm,
    ]

\addplot[
    color=cbYellow,
    mark=o,
    mark size=2pt,
    line width=1pt
] table[x=colnames, y={gemini-2.5-pro}] {\realtable};

\addplot[
    color=cbYellow,
    mark=square,
    mark size=2pt,
    line width=1pt
] table[x=colnames, y={gemini-3-pro}] {\realtable};

\addplot[
    color=cbBlue,
    mark=o,
    mark size=2pt,
    line width=1pt
] table[x=colnames, y={gpt-5-nano}] {\realtable};

\addplot[
    color=cbBlue,
    mark=x,
    mark size=2pt,
    line width=1pt
] table[x=colnames, y={gpt-5-mini}] {\realtable};

\addplot[
    color=cbBlue,
    mark=+,
    mark size=2pt,
    line width=1pt
] table[x=colnames, y={gpt-5}] {\realtable};

\addplot[
    color=cbBlue,
    mark=square,
    mark size=2pt,
    line width=1pt
] table[x=colnames, y={gpt-5.2}] {\realtable};

\addplot[
    color=cbOrange,
    mark=o,
    mark size=2pt,
    line width=1pt
] table[x=colnames, y={o4-mini}] {\realtable};

\addplot[
    color=cbOrange,
    mark=x,
    mark size=2pt,
    line width=1pt
] table[x=colnames, y={o3}] {\realtable};

\addplot[
    color=cbGreen,
    mark=o,
    mark size=2pt,
    line width=1pt
] table[x=colnames, y={claude-sonnet-4-20250514}] {\realtable};

\end{groupplot}

\end{tikzpicture}
\caption{Scores by context window length for \ours-synth and \ours-real for closed-weight models.}
\label{fig:closed_models}
\end{figure*}

\begin{figure*}[]
\centering
\begin{tikzpicture}
\pgfplotstableread[col sep=comma]{resources/plots/real_results.csv}\realdata
\pgfplotstabletranspose[colnames from=model]\realtable{\realdata}
\pgfplotstableread[col sep=comma]{resources/plots/synth_results.csv}\synthdata
\pgfplotstabletranspose[colnames from=model, columns={model,8192,16384,32768,65536,131072,262144,524288}]\synthtable{\synthdata}

\begin{groupplot}[
    group style={
        group size=2 by 1,
        horizontal sep=0.5cm,
    },
    legend style={
        at={(0.85,1.2)},
        anchor=south,
        legend columns=5,
        font=\scriptsize,
    },
    height=6cm,
]

\nextgroupplot[
    title={\ours-synth},
    xlabel={Context Length},
    ylabel={Score},
    grid=major,
    xmode=log,
    log basis x=2,
    xmin=6500, xmax=550000,
    ymin=0, ymax=1,
    xtick={8192,16384,32768,65536,131072,262144,524288},
    xticklabels={8K,16K,32K,64K,128K,256K,512K},
    width=8.5cm,
]

\addplot[
    color=cbPurple,
    mark=o,
    mark size=2pt,
    line width=1pt
] table[x=colnames, y={Llama-4-Maverick-17B-128E-Instruct-FP8}] {\synthtable};
\addlegendentry{llama-4-maverick}

\addplot[
    color=cbPurple,
    mark=square,
    mark size=2pt,
    line width=1pt
] table[x=colnames, y={Llama3-8B}] {\synthtable};
\addlegendentry{llama-3.1-8B}

\addplot[
    color=cbPurple,
    mark=x,
    mark size=2pt,
    line width=1pt
] table[x=colnames, y={Llama3-3B}] {\synthtable};
\addlegendentry{llama-3.3-3B}

\addplot[
    color=maroon,
    mark=o,
    mark size=2pt,
    line width=1pt
] table[x=colnames, y={Qwen3-4B-instruct}] {\synthtable};
\addlegendentry{Qwen3-4B-instruct}

\addplot[
    color=maroon,
    mark=x,
    mark size=2pt,
    line width=1pt
] table[x=colnames, y={Qwen3-4B-thinking}] {\synthtable};
\addlegendentry{Qwen3-4B-thinking}

\addplot[
    color=cbRed,
    mark=square,
    mark size=2pt,
    line width=1pt
] table[x=colnames, y={Olmo3.1-Think-32B}] {\synthtable};
\addlegendentry{Olmo-3.1-Think-32B}

\addplot[
    color=cbRed,
    mark=+,
    mark size=2pt,
    line width=1pt
] table[x=colnames, y={Olmo3-Think-7B}] {\synthtable};
\addlegendentry{Olmo-3-Think-7B}

\addplot[
    color=cbRed,
    mark=x,
    mark size=2pt,
    line width=1pt
] table[x=colnames, y={Olmo3-Instruct-7B}] {\synthtable};
\addlegendentry{Olmo3-Instruct-7B}

\addplot[
    color=cbRed,
    mark=o,
    mark size=2pt,
    line width=1pt
] table[x=colnames, y={Olmo3.1-Instruct-32B}] {\synthtable};
\addlegendentry{Olmo3.1-Instruct-32B}

\addplot[
    color=cbDarkYellow,
    mark=+,
    mark size=2pt,
    line width=1pt
] table[x=colnames, y={Gemma4-31B}] {\synthtable};
\addlegendentry{Gemma4-31B}

\addplot[
    color=cbGray,
     mark=o,
     mark size=2pt,
     line width=1pt
] table[x=colnames, y={deepseek-r1-0528}] {\synthtable};
\addlegendentry{deepseek-r1-0528}

\addplot[
    color=black,
    mark=.,
    dashed,
    mark size=2pt,
    line width=1pt
] table[x=colnames, y={random-baseline}] {\synthtable};
 \addlegendentry{random baseline}

\nextgroupplot[
    title={\ours-real},
    legend entries={},
    xlabel={Context Length},
    xmode=log,
    log basis x=2,
    xmin=48000, xmax=600000,
    ymin=0, ymax=1,
    xtick={32768,65536,131072,262144,524288},
    xticklabels={32K,64K,128K,256K,512K},
    yticklabel pos=right,
    grid=major,
    width=6cm,
    ]

\addplot[
    color=cbPurple,
    mark=o,
    mark size=2pt,
    line width=1pt
] table[x=colnames, y={Llama-4-Maverick-17B-128E-Instruct-FP8}] {\realtable};

\addplot[
    color=cbPurple,
    mark=square,
    mark size=2pt,
    line width=1pt
] table[x=colnames, y={Llama3-8B}] {\realtable};

\addplot[
    color=cbPurple,
    mark=x,
    mark size=2pt,
    line width=1pt
] table[x=colnames, y={Llama3-3B}] {\realtable};
%

\addplot[
    color=maroon,
    mark=o,
    mark size=2pt,
    line width=1pt
] table[x=colnames, y={Qwen3-4B-instruct}] {\realtable};

\addplot[
    color=maroon,
    mark=x,
    mark size=2pt,
    line width=1pt
] table[x=colnames, y={Qwen3-4B-thinking}]{\realtable};
%

\addplot[
    color=cbRed,
    mark=square,
    mark size=2pt,
    line width=1pt
] table[x=colnames, y={Olmo3.1-Think-32B}]{\realtable};

\addplot[
    color=cbRed,
    mark=+,
    mark size=2pt,
    line width=1pt
] table[x=colnames, y={Olmo3-Think-7B}] {\realtable};

\addplot[
    color=cbRed,
    mark=x,
    mark size=2pt,
    line width=1pt
] table[x=colnames, y={Olmo3-Instruct-7B}] {\realtable};

\addplot[
    color=cbRed,
    mark=o,
    mark size=2pt,
    line width=1pt
] table[x=colnames, y={Olmo3.1-Instruct-32B}] {\realtable};
%

\addplot[
    color=cbDarkYellow,
    mark=+,
    mark size=2pt,
    line width=1pt
] table[x=colnames, y={Gemma4-31B}] {\realtable};

\addplot[
    color=cbGray,
    mark=o,
    mark size=2pt,
    line width=1pt
] table[x=colnames, y={deepseek-r1-0528}] {\realtable};

\end{groupplot}

\end{tikzpicture}
\caption{Scores by context window length for \ours-synth and \ours-real for open-weight models.}
\label{fig:open_models}
\end{figure*}

\newpage
\section{Additional Results}
\label{app:results}

\subsection{Reasoning effort on \ours-real}

Reasoning models generally show little gain on \ours-real; the Olmo models show a small gain and the Qwen models show a small loss. When we compare ``high'' and ``low'' reasoning effort for GPT-5-nano, we see very little variation across the settings. \autoref{fig:ablation_reasoning_levels_real} shows this result for \ours-real.

\begin{figure}[ht!]
    \centering
    \begin{tikzpicture}[scale=0.8]
\begin{axis}[
    title={\ours-real},
    xlabel={Context Length},
    grid=major,
    width=7cm,
    height=6cm,
    xmode=log,
    log basis x=2,
    xmin=40000, xmax=512000,
    ymin=0, ymax=1,
    xtick={32768,65536,131072,262144,524288},
    xticklabels={32K,64K,128K,256K,512K},
]

\pgfplotstableread[col sep=comma]{resources/plots/real_results.csv}\datatable
\pgfplotstabletranspose[colnames from=model]\transposedtable{\datatable}

\addplot[
     color=cbBlue,
     mark=*,
     mark size=2pt,
     line width=1pt
 ] table[x=colnames, y={gpt-5-nano}] {\transposedtable};
 \addlegendentry{gpt-5-nano [default]}

\addplot[
    color=cbOrange,
    mark=x,
    mark size=2pt,
    line width=1pt
] table[x=colnames, y={gpt-5-nano-high}] {\transposedtable};
\addlegendentry{gpt-5-nano [high]}
 
\addplot[
    color=cbGreen,
    mark=o,
    mark size=2pt,
    line width=1pt
] table[x=colnames, y={gpt-5-nano-low}] {\transposedtable};
\addlegendentry{gpt-5-nano [low]}

\end{axis}
\end{tikzpicture}
    \caption{Comparison across reasoning levels on \ours-real.}
    \label{fig:ablation_reasoning_levels_real}
\end{figure}

\subsection{Data contamination for \ours-real}
Critical Role is an extremely popular show, with widespread discussion of the episodes online, and the statistics provided by CritRoleStats are publicly available. Could a model have simply memorized the answers to \ours-real? 

Conceptually, we believe the risk of data contamination is low: the CritRoleStats statistics are provided in spreadsheets that map from episode numbers to statistics without providing any transcript text. Since we do not provide episode numbers in our prompts, contamination could only occur if a model had memorized the statistics spreadsheet and could identify from each transcript's text the appropriate episode number. To check for potential contamination, we test the first of these two steps: whether, given the episode and campaign information without the transcript text, models can answer \ours-real questions. We use an augmented template as demonstrated below:

\begin{tcolorbox}[colback=gray!10!white, colframe=gray!50!black, boxrule=0.5pt, sharp corners]
Answer the following question about the Critical Role live-action D\&D show using ONLY your knowledge of the show-- do not look up any information. Provide your final answer in the format \textbackslash boxed\{answer\}, and attempt the question even if you are unsure. This question is about the second campaign of Critical Role: specifically, episode(s) 44. Total number of rolls in this episode?
\end{tcolorbox}

Models are unable to perform this task. We run Gemini 3.1 Pro, GPT 5.2, and Claude Sonnet 4.5 on a set of 100 random examples with this format, prompting each model 3 times. The single best run gets 6 of the 100 questions exactly correct, with Gemini 3.1 Pro answering 3.33 correctly on average, GPT 5.2 answering 0.33 on average, and Claude Sonnet 4.5 answering 1.67 correctly on average. We examined these correct answers and found that they comprised two categories: numerical answers, which we believe represent ``lucky guesses,'' as no model gave consistent numerical answers across reruns for any question; and guesses that the ``most frequent roll type'' for a span is ``attack'' or ``perception,'' which are two of the most common roll types in D\&D more broadly. We take this as evidence against widespread data contamination for \ours-real.

\section{Additional Related Work}
\label{app:related_work}

\paragraph{Dungeons \& Dragons data} Several prior works consider D\&D as a potential testbed for model capabilities; for instance, we use a version of the Critical Role transcripts preprocessed by \cite{rameshkumar-bailey-2020-storytelling}, who proposed an episode summarization task using fan-written summaries. D\&D data has also been used to study theory of mind \citep{zhou2023icastdetectthoughts}, user assistant development \cite{Zhu_Martin_Head_Callison-Burch_2023}, and  
 dialogue generation, with game state either inferred from forum-based games \citep{Callison_Burch_2022} or captured through an external tool \citep{ zhu-etal-2023-fireball}; to the best of our knowledge, we are the first to use fan annotations of gold labels and to consider the generation of these statistics as a task in its own right. 

\paragraph{Aggregation as an NLP concept} \citet{goldman-etal-2024-really} argue for the definition of long context tasks in terms of the information dispersion and input scope required to find the answer; under this taxonomy, \ours is high dispersion (because relevant info is distribution over the full context length) and high scope (because most of the input is necessary for the task, particularly for \ours-synth).
\citet{deyoung-etal-2024-multi} measure synthesis in multi-document summarization by asking if summaries convey the consensus opinion of the input documents; this is a related task to the counting tasks in \ours-synth, but framed as a measure of summarization capabilities instead of long context capabilities, and thus requires different considerations related to evaluating generated outputs. 
Relatedly, long generation benchmarks \citep{ye-2025-longproc, liu2024longgenbenchlongcontextgenerationbenchmark} measure long context ability through an entirely different lens: evaluating the ability to plan and produce coherent long outputs. 

\newpage
\clearpage
\section{ICL label validation}
\label{app:icl_label_validation}
\autoref{tab:icl_dataset_desc} describes each dataset used for \ours-synth. 
\begin{table*}[t]
\centering
\resizebox{\textwidth}{!}{
\begin{tabular}{llrr}
\toprule
Dataset  & Task & \# Labels & Input Len  \\
\midrule
Spam \citep{Almeida2011SpamFiltering} & SMS spam classification & 2 & 57\\
TREC-QC-coarse \citep{li-roth-2002-learning,hovy-etal-2001-toward} & Question type classification& 6 & 39 \\
\midrule
\midrule
AGNews \citep{Zhang2015CharacterlevelCN} & Headline topic classification & 4 & 90 \\
App Reviews \citep{Zurich-dataset} & Review sentiment classification & $2^\dagger$ & 49 \\
Pavlick Formality \citep{Lahiri-2015:arXiv,PavlickAndTetreault-2016:TACL}  & Formality classification & $2^\dagger$ &  51\\
IMDB reviews \citep{maas-EtAl:2011:ACL-HLT2011} & Sentiment analysis & $2^\dagger$ &  376\\
HiTZ Negation  \citep{garcia-ferrero-etal-2023-dataset} & Verify claims about definitions & 2& 45\\
Yahoo Topics \citep{Zhang2015CharacterlevelCN} & Question topic classification & 10 & 74 \\
MultiNLI \citep{N18-1101} &  Entailment & 3 & 70 \\
Metaphors \citep{bizzoni-lappin-2018-predicting} & Metaphor meaning validation & 2 & 51 \\
\bottomrule
\end{tabular}}
\vspace{0.2cm}
\caption{Datasets used for \ours-synth. $\dagger$ indicates cases where we combined similar labels to produce a smaller label-set for \ours. The input length is the average per-instance length (in Llama 2 tokens), including our added date and user metadata.}
\label{tab:icl_dataset_desc}

\end{table*}

\begin{table}
\centering
\begin{tabular}{lr}
\toprule
\textbf{Dataset} & \textbf{\% Removed} \\
\midrule
Spam & 0.635\% \\
TREC-QC-coarse & 0.048\% \\ 
\midrule
AGNews & 0.026\% \\
App Reviews & 0.051\% \\
Formality & 0.108\% \\
IMDB  & 0.042\% \\
Metaphors & 0.000\% \\
MultiNLI & 0.109\% \\
Negation & 0.016\% \\
Yahoo Topics & 0.188\% \\
\bottomrule
\end{tabular}

\caption{Only a small fraction of examples are removed during validation for each source dataset.}
\label{tab:validation_removal_stats}
\end{table}
 \autoref{tab:icl_dataset_desc} shows the fraction of examples removed by the validation step for each dataset.
 
 For each ICL dataset used to construct \ours-synth, we provide example instances that failed validation (and thus were discarded) and that passed validation and were used for constructing context windows. These examples are selected nearly-randomly; many of the instances that fail validation contain sexual content or offensive language, and we screen these out of the examples shown wherever possible.  While the instances screened out vary by dataset, we note that the datasets that use labels scraped from web content (i.e. Yahoo Topics, AGNews, IMDB reviews, App Reviews) generally have higher rates of validation failures. Some of these failures appear to be genuine mislabelings, which is likely because of noise in the user behavior (e.g. a user asks a question about Business \& Finance but mistakenly posts it in the Sports topic, or a user writes a review with mostly negative text but gives the product 4 stars).

\begin{table*}[htbp]
\centering
\begin{tabular}{p{8.5cm}p{2.5cm}>{\raggedleft\arraybackslash}p{2cm}}
\toprule
\textbf{Input} & \textbf{Label} & \textbf{\begin{tabular}{c}Passed\\validation?\end{tabular}} \\
\midrule
`Virtual Girlfriend' Demands Gifts HONG KONG - She needs to be coddled with sweet talk and pampered with gifts, but you'll never see her in the flesh. A Hong Kong company has developed a ``virtual girlfriend'' for new cell phones with video capability... & World & \textcolor{red}{\ding{55}} \\
\addlinespace
Court Hears Case of Brain Damaged Woman (AP) AP - The Florida Supreme Court questioned lawyers Tuesday about the extent of the power handed to Gov. Jeb Bush under a law that let him order the reinsertion of a brain-damaged woman's feeding tube. & Sci/Tech & \textcolor{red}{\ding{55}} \\
\addlinespace
Rooney backs fiercesome threesome (AFP) AFP - Wayne Rooney believes the three-pronged attack of himself, Michael Owen and Jermain Defoe can put England on the fast track to the World Cup finals. & World & \textcolor{red}{\ding{55}} \\
\addlinespace
From mouths of babes: What's hot, what's not How hot are the ``Hot Dozen'' toys? To find out, The Boston Globe put six toys from the 2004 Toy Wishes magazine Hot Dozen list in front of 13 kids from the Charlestown Boys and Girls Club. & Business & \textcolor{red}{\ding{55}} \\
\addlinespace
Crematory operator to get 12 years Ray Brent Marsh, who is to enter the plea Friday, had faced up to 8,000 years in a case that shocked the nation two years ago when investigators found hundreds of corpses at his rural northwest facility. & Business & \textcolor{red}{\ding{55}} \\
\midrule
Vending Machines Making Room for Healthy Products WASHINGTON (AP) -- The typical vending machine fare consists of chocolate bars and potato chips, leaving few options for people seeking low-calorie or low-salt snacks. That is changing now as companies develop markets for products they expect to satisfy both nutritionists and consumers... & Sci/Tech & \textcolor{green}{\ding{51}} \\
\addlinespace
Venezuelans Line Up to Vote on Chavez CARACAS, Venezuela - Summoned by bugle calls and the detonations of huge firecrackers, Venezuelans turned out in unprecedented numbers Sunday to vote on whether to force leftist President Hugo Chavez from office. Some lines at polling places extended for 1.25 miles... & World & \textcolor{green}{\ding{51}} \\
\addlinespace
Singh Snares PGA Title Vijay Singh outlasts Justin Leonard and Chris DiMarco in a three-way playoff to win the PGA Championship on Sunday at Whistling Straits in Haven, Wisconsin. & Sports & \textcolor{green}{\ding{51}} \\
\addlinespace
China's Panchen Lama visits Tibet The boy named by the Chinese authorities as the 11th Panchen Lama visits a temple in Tibet. & World & \textcolor{green}{\ding{51}} \\
\addlinespace
The Region's Highest-Paid Executives Pay for the Washington area's top executives rose significantly last year, reversing the downward trend that set in with the recession in 2001. & Business & \textcolor{green}{\ding{51}} \\
\bottomrule
\end{tabular}
\caption{AGNews examples that failed and passed validation.}

\end{table*}

\begin{table*}[htbp]
\centering
\begin{tabular}{p{8.5cm}p{2.5cm}>{\raggedleft\arraybackslash}p{2cm}}
\toprule
\textbf{Input} & \textbf{Label} & \textbf{\begin{tabular}{c}Passed\\validation?\end{tabular}} \\
\midrule
Kill Me.. & positive & \textcolor{red}{\ding{55}} \\
\addlinespace
Used to be great app. But since the last 7 updates always get - Authentication failed for: x@gmail.com & positive & \textcolor{red}{\ding{55}} \\
\addlinespace
App force closing on changing THEME PLZ FIX! & positive & \textcolor{red}{\ding{55}} \\
\addlinespace
You can't uninstall \& I don't like my space being used. But am glad for the blind \& deaf & positive & \textcolor{red}{\ding{55}} \\
\addlinespace
GOOGLE TALKBACK Read carefully \& then give at least a good reviews this app is made only for persons who got disability such as a blind person \& if you want to get rid this simply just ROOT your device dumb head!! & negative & \textcolor{red}{\ding{55}} \\
\midrule
This thing is great. I am at the bottom of the learning curve but in the few minutes I have played with it it seems like it will be easy to learn and to use. Just what I hoped for. & positive & \textcolor{green}{\ding{51}} \\
\addlinespace
Crash when set a demodulation! :( & negative & \textcolor{green}{\ding{51}} \\
\addlinespace
doesn't work with rtl-2832 Says it works with rtl devices but doesn't. never displays anything even though the rtl 2832 works with other spectrum analyzers. no help files anywhere so good luck figuring out what's wrong! & negative & \textcolor{green}{\ding{51}} \\
\addlinespace
Great app! Works really well with Galaxy note 5 & positive & \textcolor{green}{\ding{51}} \\
\addlinespace
Works great I am using HackRF One with this apps. & positive & \textcolor{green}{\ding{51}} \\
\bottomrule
\end{tabular}
\caption{App Reviews examples that failed and passed validation.}

\end{table*}

\begin{table*}[htbp]
\centering
\begin{tabular}{p{8.5cm}p{2.5cm}>{\raggedleft\arraybackslash}p{2cm}}
\toprule
\textbf{Input} & \textbf{Label} & \textbf{\begin{tabular}{c}Passed\\validation?\end{tabular}} \\
\midrule
The Unit secret agent has signed on to appear in three episodes as a businessman being shown some houses - and, presumably, one bedroom in particular - by our randy realtor heroine. & formal & \textcolor{red}{\ding{55}} \\
\addlinespace
Barely 12, with large brown eyes and stick-like arms, Fandi is 3 years older than his brother -- in his eyes almost a man. & formal & \textcolor{red}{\ding{55}} \\
\addlinespace
D-Lister Avril Lavigne appears on the cover of Z-List magazine Savvy this month. & formal & \textcolor{red}{\ding{55}} \\
\addlinespace
ORHS is going through it now, but from all I can perceive the district is out for the quick fix, chop off the principal, instead of really trying to assist the students and dig to the bottom to find the truth. & formal & \textcolor{red}{\ding{55}} \\
\addlinespace
Little Sally Draper (Kiernan Shipka) is a "Patty Duke"-era girl - indeed, "Mad Men" is currently set during the month "The Patty Duke Show" premiered in 1963 - and on a "Patty Duke"-like show, she'd just be a lispy Shirley Temple doll with a crush on Daddy. & formal & \textcolor{red}{\ding{55}} \\
\midrule
I have tried everything possible to attract business. & formal & \textcolor{green}{\ding{51}} \\
\addlinespace
Never heard that one before. & informal & \textcolor{green}{\ding{51}} \\
\addlinespace
"Yup." & informal & \textcolor{green}{\ding{51}} \\
\addlinespace
That's how we pass our traits to the next generation: through DNA & formal & \textcolor{green}{\ding{51}} \\
\addlinespace
Wedding rituals differ in different regions and communities in India. & formal & \textcolor{green}{\ding{51}} \\
\bottomrule
\end{tabular}
\caption{Formality examples that failed and passed validation.}

\end{table*}
\begin{table*}[t]
\centering
\resizebox{0.95\textwidth}{!}{
\begin{tabular}{p{10cm}p{1.5cm}>{\raggedleft\arraybackslash}p{2cm}}
\toprule
\textbf{Input} & \textbf{Label} & \textbf{\begin{tabular}{c}Passed\\validation?\end{tabular}} \\
\midrule
Good movie, very 70s, you can not expect much from a film like this. Sirpa Lane is an actress of erotic films, a nice body but nothing exceptional. Not demand a lot from these films are light years away from the movies today, the world has changed incredibly. The plot is simple and the actors not extraordinary. & positive & \textcolor{red}{\ding{55}} \\
\addlinespace
This film Evil Breed: The legend of samhain contains very little thought or effort. It is ridiculed with specs of ultra fast "slasher" style death and plain disgusting acts of death. The acting was rated a D as the actors show very little ability, and the stupidity of them in the film is too questionable. & positive & \textcolor{red}{\ding{55}} \\
\addlinespace
The movie "Holly" may make the audience want to donate money towards organizations that improve the life for these poor youngsters, but the film's dramatic weaknesses may reduce its chances of being seen by enough people to make a difference. Overall, I think the concept is better as a documentary and it was not as touching as a movie. & positive & \textcolor{red}{\ding{55}} \\
\addlinespace
Although this film put Davis on the map due to her brilliantly intense performance, this film is strangely unsatisfying to me as a whole. What I cannot fathom for the life of me is just how or why Phillip would take the constant abuse this tramp constantly dishes out towards him. & positive & \textcolor{red}{\ding{55}} \\
\midrule
This film is just plain horrible. John Ritter doing pratt falls, 75\% of the actors delivering their lines as if they were reading them from cue cards, poor editing, horrible sound mixing, and a plot that really goes nowhere. If I could sum this film up in one word, that word would be: Suckotrocity & negative & \textcolor{green}{\ding{51}} \\
\addlinespace
Zentropa has much in common with The Third Man, another noir-like film set among the rubble of postwar Europe. Like TTM, there is much inventive camera work. There is an innocent American who gets emotionally involved with a woman he doesn't really understand, and whose naivety is all the more striking in contrast with the natives. But I'd have to say that The Third Man has a more well-crafted storyline. Zentropa is a bit disjointed in this respect. Perhaps this is intentional: it is presented as a dream/nightmare, and making it too coherent would spoil the effect. This movie is unrelentingly grim--``noir'' in more than one sense; one never sees the sun shine. Grim, but intriguing, and frightening. & positive & \textcolor{green}{\ding{51}} \\
\addlinespace
Never in my life have I come across a movie as bad as The Zombie Chronicles. Filmed on a budget of what looks to be about 20 bucks, TZC is a completely horrible horror movie that relies on lame, forgettable actors. Simply put, avoid TZC like a sexually-transmitted disease. & negative & \textcolor{green}{\ding{51}} \\
\addlinespace
Without wishing to be a killjoy, Brad Sykes is responsible for at least two of the most dull and clichéd films I've ever seen. The acting is terrible, the print is shoddy, and everything about this film screams "seriously, you could do better yourself". & negative & \textcolor{green}{\ding{51}} \\
\bottomrule
\end{tabular}}
\caption{IMDB examples that failed and passed validation. Only eight examples are shown because of the example length.}

\end{table*}
\begin{table*}[htbp]
\centering
\begin{tabular}{p{8.5cm}p{2.5cm}>{\raggedleft\arraybackslash}p{2cm}}
\toprule
\textbf{Input} & \textbf{Label} & \textbf{\begin{tabular}{c}Passed\\validation?\end{tabular}} \\
\midrule
Draupadi's eyes were diamonds. $\leftrightarrow$ Draupadi's eyes were beautiful. & correct & \textcolor{green}{\ding{51}} \\
\addlinespace
The faculty meeting was an easy breeze $\leftrightarrow$ The faculty meeting was very easy and relaxing & correct & \textcolor{green}{\ding{51}} \\
\addlinespace
The house was a tomb. $\leftrightarrow$ The house was big. & incorrect & \textcolor{green}{\ding{51}} \\
\addlinespace
I had already planted the idea in her mind. $\leftrightarrow$ I had already scared her about the idea. & incorrect & \textcolor{green}{\ding{51}} \\
\addlinespace
It is sad to observe the fruits of ignorance. $\leftrightarrow$ It is sad to observe the effects of ignorance. & correct & \textcolor{green}{\ding{51}} \\
\bottomrule
\end{tabular}
\caption{Metaphors examples that passed validation. No examples failed validation for this dataset.}

\end{table*}
\begin{table*}[htbp]
\centering
\begin{tabular}{p{8.5cm}p{2.5cm}>{\raggedleft\arraybackslash}p{2cm}}
\toprule
\textbf{Input} & \textbf{Label} & \textbf{\begin{tabular}{c}Passed\\validation?\end{tabular}} \\
\midrule

You're safe. $\rightarrow$ You have nothing to worry about. & neutral & \textcolor{red}{\ding{55}} \\
\addlinespace
asks Burton, cupping his ear. $\rightarrow$ Burton wanted to ask what was going on, but he couldn't do much of anything while bound and gagged. & contradiction & \textcolor{red}{\ding{55}} \\
\addlinespace
I ``poison my dearest Emily!'' $\rightarrow$ How could you think I would poison Emily? & entailment & \textcolor{red}{\ding{55}} \\
\addlinespace
When he finally succeeded, after a prolonged siege and heavy losses, he punished the local population by cutting off the noses and lips of all men except those who played wind instruments. $\rightarrow$ All men who played wind instruments were tasked with helping cut off people's noses. & neutral & \textcolor{red}{\ding{55}} \\
\addlinespace
They died slowly, their eyes bulging and faces turning blue. $\rightarrow$ They died asphyxiated. & neutral & \textcolor{red}{\ding{55}} \\
\midrule
well you see that on television also $\rightarrow$ You can see that on television, as well. & entailment & \textcolor{green}{\ding{51}} \\
\addlinespace
Vrenna and I both fought him and he nearly took us. $\rightarrow$ Neither Vrenna nor myself have ever fought him. & contradiction & \textcolor{green}{\ding{51}} \\
\addlinespace
This analysis pooled estimates from these two studies to develop a C-R function linking PM to chronic bronchitis. $\rightarrow$ The analysis proves that there is no link between PM and bronchitis. & contradiction & \textcolor{green}{\ding{51}} \\
\addlinespace
He turned and smiled at Vrenna. $\rightarrow$ He smiled at Vrenna who was walking slowly behind him with her mother. & neutral & \textcolor{green}{\ding{51}} \\
\addlinespace
We sought to identify practices that were commonly implemented by the agencies within the past 5 years. $\rightarrow$ We want to identify practices commonly used by agencies in the last 5 years & entailment & \textcolor{green}{\ding{51}} \\

\bottomrule
\end{tabular}
\caption{MultiNLI examples that failed and passed validation.}

\end{table*}
\begin{table*}[htbp]
\centering
\begin{tabular}{p{8.5cm}p{2.5cm}>{\raggedleft\arraybackslash}p{2cm}}
\toprule
\textbf{Input} & \textbf{Label} & \textbf{\begin{tabular}{c}Passed\\validation?\end{tabular}} \\
\midrule
No care is anything that serves as an enticement. & True & \textcolor{red}{\ding{55}} \\
\addlinespace
No care may be anything that serves as an enticement. & True & \textcolor{red}{\ding{55}} \\
\addlinespace
A loud utterance is an appropriate definition of produce in no context. & True & \textcolor{red}{\ding{55}} \\
\addlinespace
Complete attention is an appropriate definition of candidate in no context. & True & \textcolor{red}{\ding{55}} \\
\addlinespace
No stress is a message received and understood. & True & \textcolor{red}{\ding{55}} \\
\midrule
Action refers to a military engagement. & True & \textcolor{green}{\ding{51}} \\
\addlinespace
Action never stands for a distinguishing quality. & True & \textcolor{green}{\ding{51}} \\
\addlinespace
Action does not stand for a military engagement. & False & \textcolor{green}{\ding{51}} \\
\addlinespace
Action may stand for a military engagement. & True & \textcolor{green}{\ding{51}} \\
\addlinespace
Not a single action is a military engagement. & False & \textcolor{green}{\ding{51}} \\

\bottomrule
\end{tabular}
\caption{Negation examples that failed and passed validation.}

\end{table*}
\begin{table*}[htbp]
\centering
\begin{tabular}{p{8.5cm}p{2.5cm}>{\raggedleft\arraybackslash}p{2cm}}
\toprule
\textbf{Input} & \textbf{Label} & \textbf{\begin{tabular}{c}Passed\\validation?\end{tabular}} \\
\midrule
Go chase after her and run her over while she's crossing the street & ham & \textcolor{red}{\ding{55}} \\
\addlinespace
i want to grasp your pretty booty :) & ham & \textcolor{red}{\ding{55}} \\
\addlinespace
No da if you run that it activate the full version da. & ham & \textcolor{red}{\ding{55}} \\
\addlinespace
i am seeking a lady in the street and a freak in the sheets. Is that you? & ham & \textcolor{red}{\ding{55}} \\
\addlinespace
Should i send you naughty pix? :) & ham & \textcolor{red}{\ding{55}} \\
\midrule
WINNER!! As a valued network customer you have been selected to receive a £900 prize reward! To claim call 09061701461. Claim code KL341. Valid 12 hours only. & spam & \textcolor{green}{\ding{51}} \\
\addlinespace
Had your mobile 11 months or more? U R entitled to Update to the latest colour mobiles with camera for Free! Call The Mobile Update Co FREE on 08002986030 & spam & \textcolor{green}{\ding{51}} \\
\addlinespace
I'm gonna be home soon and i don't want to talk about this stuff anymore tonight, k? I've cried enough today. & ham & \textcolor{green}{\ding{51}} \\
\addlinespace
SIX chances to win CASH! From 100 to 20,000 pounds txt> CSH11 and send to 87575. Cost 150p/day, 6days, 16+ TsandCs apply Reply HL 4 info & spam & \textcolor{green}{\ding{51}} \\
\addlinespace
I HAVE A DATE ON SUNDAY WITH WILL!! & ham & \textcolor{green}{\ding{51}} \\
\bottomrule
\end{tabular}
\caption{Spam examples that failed and passed validation. The examples that failed validation were almost exclusively ham messages with sexual content; the selected messages are the least inappropriate of the examples that failed validation.}

\end{table*}
\begin{table*}[htbp]
\centering
\begin{tabular}{p{8.5cm}p{2.5cm}>{\raggedleft\arraybackslash}p{2cm}}
\toprule
\textbf{Input} & \textbf{Label} & \textbf{\begin{tabular}{c}Passed\\validation?\end{tabular}} \\
\midrule
What explosive do you get by mixing charcoal, sulfur and saltpeter? & entity & \textcolor{red}{\ding{55}} \\
\addlinespace
What is the procedure called for drilling a hole in your skull to achieve a higher consciousness? & entity & \textcolor{red}{\ding{55}} \\
\midrule
Name 11 famous martyrs. & human being & \textcolor{green}{\ding{51}} \\
\addlinespace
What's the Olympic motto? & description and abstract concept & \textcolor{green}{\ding{51}} \\
\addlinespace
What is the highest waterfall in the United States? & location & \textcolor{green}{\ding{51}} \\
\addlinespace
What does the abbreviation AIDS stand for? & abbreviation & \textcolor{green}{\ding{51}} \\
\addlinespace
How many points make up a perfect fivepin bowling score? & numeric value & \textcolor{green}{\ding{51}} \\
\bottomrule
\end{tabular}
\caption{TREC-coarse-Q examples that failed and passed validation. Only two examples failed validation.}

\end{table*}
\begin{table*}[htbp]
\centering
\begin{tabular}{p{8.5cm}p{2.5cm}>{\raggedleft\arraybackslash}p{2cm}}
\toprule
\textbf{Input} & \textbf{Label} & \textbf{\begin{tabular}{c}Passed\\validation?\end{tabular}} \\
\midrule
Is there a God? The question to end all questions, and begin them. & Business \& Finance & \textcolor{red}{\ding{55}} \\
\addlinespace
why do we need to lie? & Health & \textcolor{red}{\ding{55}} \\
\addlinespace
try to type the word supercalifragilisticexpialidocious. 20 times fast you can only make 10 mistakes? This is for fun! & Education \& Reference & \textcolor{red}{\ding{55}} \\
\addlinespace
do you think it is okay to tell a lie? & Education \& Reference & \textcolor{red}{\ding{55}} \\
\addlinespace
where can you purchase cesium carbonate? & Science \& Mathematics & \textcolor{red}{\ding{55}} \\
\midrule
What are good sources to find out about new gospel artists? Is there a site that focuses primarily on gospel? & Entertainment \& Music & \textcolor{green}{\ding{51}} \\
\addlinespace
How a black hole is formed? I would like to know how a black hole can possibly be formed. Are there any experimental evidence of such creation? & Science \& Mathematics & \textcolor{green}{\ding{51}} \\
\addlinespace
Economics of running a restaurant? Running a restaurant looks like hard work and long hours. What percentage of restaurants are profitable? & Business \& Finance & \textcolor{green}{\ding{51}} \\
\addlinespace
Why doesn't the NBA implement a minor leagues? I don't want to see any more High School kids on the court. & Sports & \textcolor{green}{\ding{51}} \\
\addlinespace
how does a helicopter fly this is a miracle- I always wanted to learn to fly one of those. Can someone explain how can it get airborne? & Science \& Mathematics & \textcolor{green}{\ding{51}} \\
\bottomrule
\end{tabular}
\caption{Yahoo examples that failed and passed validation. One class of examples that failed validation, not demonstrated here, are sexually explicit questions labeled as seemingly random categories (likely from spam posters on the original Yahoo Answers forum).}

\end{table*}

\newpage
\section{Question types}

\subsection{\ours-synth}
\label{app:app_oolong_synth}

\autoref{tab:oolong_synth_counting}, \autoref{tab:oolong_synth_user}, and \autoref{tab:oolong_synth_timeline} list the questions used for each type of task in \ours-synth.
\begin{table*}
\centering
\begin{tabular}{p{0.95\textwidth}}
\toprule
Counting \\
\midrule
In the above data, which of the labels is the most common? Give your final answer in the form `label: answer' where answer is one of the labels: \{label\_list\}. \\
\addlinespace
In the above data, which of the labels is the least common? Give your final answer in the form `label: answer' where answer is one of the labels: \{label\_list\}. \\
\addlinespace
In the above data, is label `\{A\}' more common, less common, or the same frequency as label `\{B\}'? Give your final answer in the form `Answer: \{A\} is \lbrack X\rbrack  \{B\}', where \lbrack X\rbrack  is `more common than', `less common than', or 'same frequency as'.\\
\addlinespace
In the above data, how many data points should be classified as label `\{label\}'? Give your final answer in the form `Answer: number'. \\

\bottomrule
\end{tabular}
\caption{Counting questions used in the \ours-synth dataset.}
\label{tab:oolong_synth_counting}
\end{table*}

\begin{table*}
\centering
\begin{tabular}{p{0.95\textwidth}}
\toprule
User \\
\midrule

In the above data, which user is represented most often? Give your final answer in the form `User: \lbrack X\rbrack ', where \lbrack X\rbrack  is the user ID.\\
\addlinespace
In the above data, which user is represented the second most often? Give your final answer in the form `User: \lbrack X\rbrack ', where \lbrack X\rbrack  is the user ID.\\
\addlinespace
For the following question, only consider the subset of users with IDs \{user\_names\}. Among these users, which user is represented most often? Give your final answer in the form `User: \lbrack X\rbrack ', where \lbrack X\rbrack  is the user ID.\\
\addlinespace
For the following question, only consider the subset of users with IDs \{user\_names\}. Among these users, which user is represented the second most often? Give your final answer in the form `User: \lbrack X\rbrack ', where \lbrack X\rbrack  is the user ID.\\
\addlinespace
For the following question, only consider the subset of instances that are associated with user IDs \{user\_names\}. Among instances associated with these users, \{any of the Counting questions above\}  \\
\addlinespace
For the following question, only consider the subset of users with IDs \{user\_names\}. Among these users, which user has the most instances with the label \{label\}? Give your final answer in the form `User: \lbrack X\rbrack ', where \lbrack X\rbrack  is the user ID. \\
\addlinespace
In the above data, which user has the most instances with the label \{label\}? Give your final answer in the form `User: \lbrack X\rbrack ', where \lbrack X\rbrack  is the user ID. \\
\addlinespace
In the above data, which user has more instances with the label \{label\}: User \{A\} or User \{B\}? Give your final answer in the form `User: \lbrack X\rbrack ', where \lbrack X\rbrack  is the user ID.\\

\bottomrule
\end{tabular}
\caption{User questions used in the \ours-synth dataset.}
\label{tab:oolong_synth_user}
\end{table*}

\begin{table*}
\centering
\begin{tabular}{p{0.95\textwidth}}
\toprule
Timeline \\
\midrule
In the above data, which date is represented most often? Give your final answer in the form `Date: \lbrack X\rbrack ', where \lbrack X\rbrack  is the date in the format MM/DD/YYYY. \\
\addlinespace
In the above data, which date is represented most often? Give your final answer in the form `Date: \lbrack X\rbrack ', where \lbrack X\rbrack  is the date in the format MM/DD/YYYY. \\
\addlinespace
In the above data, which date is represented the second most often? Give your final answer in the form `Date: \lbrack X\rbrack ', where \lbrack X\rbrack  is the date in the format MM/DD/YYYY. \\
\addlinespace
In the above data, how many dates are represented exactly \{n\} times? Give your final answer in the form `Answer: \lbrack X\rbrack ', where \lbrack X\rbrack  is the number of dates represented exactly \{n\} times.\\
\addlinespace
In the above data, was label `\{key\}' more common, less common, or the same frequency before \{time\}, as compared to after \{time\}? Give your final answer in the form `Answer: \{key\} is \lbrack X\rbrack  before \{time\}', where \lbrack X\rbrack  is `more common', `less common', or 'the same frequency'.\\
\addlinespace
In the above data, was label `\{key\}' more common, less common, or the same frequency before \{time\}, as compared to after \{time\}? Give your final answer in the form `Answer: \{key\} is \lbrack X\rbrack  before \{time\}', where \lbrack X\rbrack  is `more common', `less common', or 'the same frequency'.\\
\addlinespace
For the following question, only consider the subset of instances that occur in \{month\_name\} of any year. Among instances occuring in \{month\_name\},\{any of the Counting questions above\}  \\
\addlinespace
For the following question, only consider the subset of instances that occur between \{starting\_date\} and \{ending\_date\}, inclusive. Among instances occuring in this date range, \{any of the Counting questions above\} \\
\addlinespace
In which month did the label `\{label1\} first occur more often than the label `\{label2\}'? Give your final answer in the form `Answer: \lbrack month\rbrack  \lbrack year\rbrack ', where \lbrack month\rbrack  is the name of the month and \lbrack year\rbrack  is the four-digit year where `\{label1\}' first occured more often than `\{label2\}.'\\
\addlinespace
For how many months does the label `\{label1\}' occur more frequently than the label `\{label2\}'? Disregard months where there is a tie.  Give your final answer in the form `Answer: \lbrack X\rbrack ', where \lbrack X\rbrack  is the number of months where `\{label1\}' occurs more often than `\{label2\}.' \\
\addlinespace
For how many months is the label `\{label\}' the single most frequently occuring label? Disregard months where there is a tie for the most common label.  Give your final answer in the form `Answer: \lbrack X\rbrack ', where \lbrack X\rbrack  is the number of months where `\{label\}' is the most common label. \\

\bottomrule
\end{tabular}
\caption{Timeline questions used in the \ours-synth dataset.}
\label{tab:oolong_synth_timeline}
\end{table*}

\subsection{\ours-real}
\label{ssec:app_oolong_real}

\autoref{tab:oolong_real_single} lists the question types used for single episode settings in \ours-real dataset. In \autoref{tab:oolong_real_multi_rolls}, \autoref{tab:oolong_real_multi_spells} lists the types for multi-episode settings.

\begin{table*}[t]
\centering
\begin{tabular}{p{0.95\textwidth}}
\toprule
Rolls \\
\midrule
Total number of rolls in this episode? (counting) \\
Total number of rolls by the character \{character name\} in this episode? (counting, character) \\
Total number of rolls by the player \{player name\} in this episode? (counting, player) \\
Total number of rolls of type \{roll type\} in this episode? (counting, roll type) \\
Number of rolls of natural value \{roll value\} in this episode? (counting, roll value) \\
In this episode, what percentage of rolls were of value \{roll value\}? round to the nearest integer. (counting, roll value) \\
What is the most common roll type in this episode? Return a comma separated list. (frequency) \\
What is the least common roll type in this episode? Only include types with more than one roll. Return a comma separated list. (frequency) \\
What is the most common natural roll value in this episode? Return a comma separated list. (frequency) \\
What is the least common natural roll value in this episode? Only include values with more than one roll. Return a comma separated list. (frequency) \\
What is the count of Crits? (natural rolls of value 1 or 20)? (counting, roll value) \\
What is the count of Nat20s (natural rolls of value 20)? (counting, roll value) \\
What is the count of Nat1s (natural rolls of value 1)? (counting, roll value) \\
\midrule
Spells \\
\midrule
How many spells were cast during this episode? (counting) \\
How many spells were cast by the character \{character name\} in this episode? (counting, character) \\
How many spells were cast by the player \{player name\} in this episode? (counting, player) \\
How many \{spell type\} spells were cast during this episode? (counting, spell type) \\
What is the first spell cast in this episode? (enumeration) \\
What are the first two spells cast in this episode? (enumeration) \\
What are the first three spells cast in this episode? (enumeration) \\
What is the last spell cast in this episode? (enumeration) \\
What are the last two spells cast in this episode? (enumeration) \\
What are the last three spells cast in this episode? (enumeration) \\
What is the first spell cast by each character in this episode? Return a comma separated list and retain the order of spells as they appear in the episode. (enumeration) \\
What is the last spell cast by each character in this episode? Return a comma separated list and retain the order of spells as they appear in the episode. (enumeration) \\
How many characters cast \{spell name\} spell in this episode? (counting, spell type) \\
What is the most common spell in this episode? Return a comma separated list. (frequency) \\
What is the least common spell in this episode? Only include spells that were cast at least once. Return a comma separated list. (frequency) \\
Which spells were cast by more than one character in this episode? Return a comma separated list. (counting, enumeration) \\
What is the total number of cantrip spells cast in this episode? (counting, spell type) \\
In this episode, how many times was a spell cast at a level higher than its base level? (counting) \\
In this episode, which spells were cast at a level higher than their base level? Return a comma separated list of unique spells. (enumeration) \\
\bottomrule
\end{tabular}
\caption{Question types used in the \ours-real dataset (single episode). For reference, we provide the question types in parentheses.}
\label{tab:oolong_real_single}
\end{table*}
\begin{table*}[t]
\centering
\begin{tabular}{p{0.95\textwidth}}
\toprule
Rolls \\
\midrule
What is the cumulative total of rolls by the end of episode \{episode index\}? Count the number of rolls and not the values of the rolls. (counting) \\
What is the cumulative total of rolls by the character \{character name\} at the end of episode \{episode index\}? Count the number of rolls and not the values of the rolls. (counting, character) \\
Total number of rolls across all the episodes? (counting) \\
Total number of rolls by the character \{character name\} across all episodes? (counting, character) \\
Total number of rolls by the player \{player name\} across all episodes? (counting, player) \\
Total number of rolls of type \{roll type\} across all episodes? (counting, roll type) \\
Number of rolls of natural value \{roll value\} across all episodes? (counting, roll value) \\
Across all episodes, what percentage of rolls were of value \{roll value\}? round to the nearest integer. (counting, roll value) \\
What is the most common roll type across all episodes? Return a comma separated list. (frequency) \\
What is the least common roll type across all episodes? Only include types with more than one roll. Return a comma separated list. (frequency) \\
What is the most common natural roll value across all episodes? Return a comma separated list. (frequency) \\
What is the least common natural roll value across all episodes? Only include values with more than one roll. Return a comma separated list. (frequency) \\
What is the total count of Crits across all episodes? (natural rolls of value 1 or 20)? (counting, roll value) \\
What is the total count of Nat20s across all episodes? (natural rolls of value 20)? (counting, roll value) \\
What is the total count of Nat1s across all episodes? (natural rolls of value 1)? (counting, roll value) \\
\bottomrule
\end{tabular}
\caption{Question types used in the \ours-real dataset (multi-episode). For reference, we provide the question types in parentheses.}
\label{tab:oolong_real_multi_rolls}
\end{table*}

\begin{table*}[t]
\centering
\begin{tabular}{p{0.95\textwidth}}
\toprule
Spells \\
\midrule
What is the cumulative total of spells cast by the end of episode \{episode index\}? (counting, indexing) \\
What is the first spell cast in the episode \{episode index\}? (enumeration, indexing) \\
What is the second spell cast in the episode \{episode index\}? (enumeration, indexing) \\
What is the third spell cast in the episode \{episode index\}? (enumeration, indexing) \\
List the first spell cast in each episode? Return a comma separated list. (enumeration) \\
List the last spell cast in each episode? Return a comma separated list. (enumeration) \\
List the first spell cast by the character \{character name\} in each episode? Return a comma separated list. (enumeration, character) \\
List the last spell cast by the character \{character name\} in each episode? Return a comma separated list. (enumeration, character) \\
How many spells were cast across all episodes? (counting) \\
How many spells were cast by the character \{character name\} across all episodes? (counting, character) \\
How many spells were cast by the player \{player name\} across all episodes? (counting, player) \\
How many \{spell type\} spells were cast across all episodes? (counting, spell type) \\
How many characters cast \{spell name\} spell across all episodes? (counting, spell type) \\
What is the most common spell across all episodes? Return a comma separated list. (frequency, enumeration) \\
What is the least common spell across all episodes? Only include spells that were cast at least once. Return a comma separated list. (frequency, enumeration) \\
What is the total number of cantrip spells cast across all episodes? (counting, spell type) \\
Across all episodes, how many times was a spell cast at a level higher than its base level? (counting, spell level) \\
Across all episodes, which spells were cast at a level higher than their base level? Return a comma separated list of unique spells. (enumeration, spell level) \\
\bottomrule
\end{tabular}
\caption{Question types used in the \ours-real dataset (multi-episode). For reference, we provide the question types in parentheses.}
\label{tab:oolong_real_multi_spells}
\end{table*}

\newpage
\section{Deepseek R1 trace examples}
\label{app:deepseek_example_traces}
\autoref{fig:deepseek_synth_trace} and \autoref{fig:deepseek_real_trace} contrast the behavior of this model on representative examples from the \ours-synth and \ours-real splits. See the main text for more discussion. 
\newpage
\clearpage
\onecolumn
\begin{tcolorbox}[colback=gray!10!white, colframe=gray!50!black, boxrule=0.5pt, sharp corners]
$<$think$>$ We are given 21 pairs. We need to classify each pair as one of `neutral', `contradiction', or `entailment'. Then, we will count the frequency of 'contradiction' and 'entailment' and compare them. Let's go through each pair one by one: 1. ``and then the class uh double A Dallas Fort Worth Spurs we later uh they went to the Texas league - Dallas Fort Worth Spurs got into the Texas League.''   - The first sentence says they went to the Texas league. The second says they got into the Texas League. This seems to be the same meaning. So, entailment. 2. ``Specifically, although most public companies are required to have a board that is comprised of a majority of independent directors and certain key committees are required to be comprised solely of outside directors, a closer look reveals that the independence of many boards may be more a matter of form than substance. - Public companies can put whoever they want on their boards.''  - The first sentence says that public companies are required to have independent directors, but the independence might be in form only. The second sentence says they can put whoever they want. This contradicts the requirement mentioned. So, contradiction. \textbf{[....authors truncated  classifications for space] }19. ``And the bureaucratic approach prevents the section as a whole from taking positions on--or creating--literary issues. - Bureaucracy makes groups not take positions or a stand on literary issues.''   - The second is a restatement of the first. Entailment. 20. ``They are linked not only by geographical location but also by here are the upmarket and fashionable neighborhoods, with affluent residential sections, trendy restaurants, and L.A.'s fabulous shopping areas. - They are linked because they both have very exclusive shoe stores.''    - The first says they are linked by upmarket and fashionable neighborhoods, etc. The second says they are linked because of exclusive shoe stores. The first does not mention shoe stores. So, the second is a specific reason that is not stated. This could be neutral. However, note that the first lists shopping areas, which might include shoe stores, but it doesn't specify. The second is too
\captionof{figure}{Example reasoning trace for \ours-synth. The model fails to complete the task because it runs out of tokens during the reasoning trace. The majority of the trace (classifying items 3-18) is omitted for space.}
\label{fig:deepseek_synth_trace}
\end{tcolorbox}

\newpage
\clearpage
\begin{tcolorbox}[colback=gray!10!white, colframe=gray!50!black, boxrule=0.5pt, sharp corners]
Based on the provided transcript of Episode 2, the second spell cast is Disintegrate, cast by Lyra (Felicia Day's character).

Sequence of Events:
1. First Spell: Vex'ahlia (Laura Bailey) casts Hunter's Mark on a frost giant (during her turn in combat).
2. Second Spell: Lyra (Felicia Day) casts Disintegrate on the same frost giant (during her subsequent turn).

Key Excerpt:
Felicia: ``I have a spell called Disintegrate. Can I cast that on its wing? [...] I'm doing Disintegrate. And it's 10d6 plus a d10, plus 40.''
Matt: ``Disintegrate is nasty. [...] As you release the torrent of electrical energy, it courses through the body [...] the giant looks severely messed up.''

Thus, the second spell in Episode 2 is Disintegrate.

Disintegrate
\captionof{figure}{Example reasoning trace for \ours-real. The model completes the task within a relatively short reasoning trace, despite this task using substantially longer input.}
\label{fig:deepseek_real_trace}
\end{tcolorbox}
\twocolumn

\newpage
\onecolumn
\section{Question type analysis }
\label{app:question_type_analysis}
\subsection{\ours-synth}

On \ours-synth, we further break down results by question type (counting, user, or temporal) and answer type (the expected format of the answer, e.g. as a label or number or date) in \autoref{fig:answer_type}. Consistently, temporal questions are the most challenging for models, highlighting the difficulty of temporal reasoning. \autoref{fig:answer_type} shows that temporal questions are challenging for models; this is reflected in both the performance by task type and the distribution of score by answer type. Questions that require a date or month/year (e.g. ``January 2021'') as an answer show generally lower performance for the same model, and show greater spread in model capabilities than the other answer categories. For instance, the gap between GPT-5 and GPT-5-nano performance is more than 4x larger for questions that require outputting a date than for questions that require outputting a label. Relative model performance is mostly stable across answer types, although Claude-Sonnet-4 is relatively much stronger on numerical reasoning and comparisons than the GPT series models.

\begin{figure*}[ht]
\centering
\definecolor{geminicolor}{RGB}{230,190,60}      
\definecolor{gpt5nanocolor}{RGB}{30,60,120}     
\definecolor{gpt5minicolor}{RGB}{70,130,200}    
\definecolor{gpt5color}{RGB}{70,130,200}        
\definecolor{o4minicolor}{RGB}{220,150,50}      
\definecolor{o3color}{RGB}{230,180,50}          
\definecolor{llamacolor}{RGB}{100,200,220}      
\definecolor{claudecolor}{RGB}{60,160,80}       
\definecolor{deepseekcolor}{RGB}{200,100,180}   

\begin{tikzpicture}[scale=0.8]

\pgfplotsset{
    mystyle/.style={
        height=7cm,
        ymin=0, ymax=1,
        ymajorgrids=true,
        grid style={dashed, gray!40},
        ytick={0,0.2,0.4,0.6,0.8,1},
        ylabel={Score},
        ylabel style={font=\small},
        tick label style={font=\small},
    }
}

\begin{axis}[
    mystyle,
    width=12cm,
    name=left,
    xmin=0.5, xmax=6.5,
    xtick={1,2,3,4,5,6},
    xticklabels={label, comparison, user, number, month+year, date},
    xticklabel style={font=\small},
    xlabel={Answer Type},
    xlabel style={font=\small},
]

\addplot[only marks, mark=*, mark size=3pt, geminicolor] 
    table[col sep=comma, x=x, y=y]{resources/plots/synth_analysis/answer_gemini-2.5-pro.csv};

\addplot[only marks, mark=*, mark size=3pt, gpt5nanocolor] 
    table[col sep=comma, x=x, y=y]{resources/plots/synth_analysis/answer_gpt-5-nano.csv};

\addplot[only marks, mark=x, mark size=4pt, gpt5minicolor, very thick] 
    table[col sep=comma, x=x, y=y]{resources/plots/synth_analysis/answer_gpt-5-mini.csv};

\addplot[only marks, mark=-, mark size=5pt, gpt5color, line width=2pt] 
    table[col sep=comma, x=x, y=y]{resources/plots/synth_analysis/answer_gpt-5.csv};

\addplot[only marks, mark=*, mark size=3pt, o4minicolor] 
    table[col sep=comma, x=x, y=y]{resources/plots/synth_analysis/answer_o4-mini.csv};

\addplot[only marks, mark=+, mark size=4pt, o3color, very thick] 
    table[col sep=comma, x=x, y=y]{resources/plots/synth_analysis/answer_o3.csv};

\addplot[only marks, mark=*, mark size=3pt, llamacolor] 
    table[col sep=comma, x=x, y=y]{resources/plots/synth_analysis/answer_Llama-4-Maverick-17B-128E-Instruct-FP8.csv};

\addplot[only marks, mark=*, mark size=3pt, claudecolor] 
    table[col sep=comma, x=x, y=y]{resources/plots/synth_analysis/answer_claude-sonnet-4-20250514.csv};

\addplot[only marks, mark=*, mark size=3pt, deepseekcolor] 
    table[col sep=comma, x=x, y=y]{resources/plots/synth_analysis/answer_deepseek-r1-0528.csv};

\addplot[only marks, mark=square*, mark size=3pt, gpt5nanocolor] 
    table[col sep=comma, x=x, y=y]{resources/plots/synth_analysis/answer_gpt-5.2.csv};

\addplot[only marks, mark=square*, mark size=3pt, geminicolor] 
    table[col sep=comma, x=x, y=y]{resources/plots/synth_analysis/answer_gemini-3-pro-preview.csv};

\end{axis}

\begin{axis}[
    mystyle,
    width=6cm,
    name=right,
    at={(left.east)},
    anchor=west,
    xshift=1cm,
    xmin=0.5, xmax=3.5,
    xtick={1,2,3},
    xticklabels={counting, user, timeline},
    xticklabel style={font=\small},
    xlabel={Task Type},
    xlabel style={font=\small},
    yticklabel pos=right,
    ylabel={},
]

\addplot[only marks, mark=*, mark size=3pt, geminicolor] 
    table[col sep=comma, x=x, y=y]{resources/plots/synth_analysis/task_gemini-2.5-pro.csv};

\addplot[only marks, mark=*, mark size=3pt, gpt5nanocolor] 
    table[col sep=comma, x=x, y=y]{resources/plots/synth_analysis/task_gpt-5-nano.csv};

\addplot[only marks, mark=x, mark size=4pt, gpt5minicolor, very thick] 
    table[col sep=comma, x=x, y=y]{resources/plots/synth_analysis/task_gpt-5-mini.csv};

\addplot[only marks, mark=-, mark size=5pt, gpt5color, line width=2pt] 
    table[col sep=comma, x=x, y=y]{resources/plots/synth_analysis/task_gpt-5.csv};

\addplot[only marks, mark=*, mark size=3pt, o4minicolor] 
    table[col sep=comma, x=x, y=y]{resources/plots/synth_analysis/task_o4-mini.csv};

\addplot[only marks, mark=+, mark size=4pt, o3color, very thick] 
    table[col sep=comma, x=x, y=y]{resources/plots/synth_analysis/task_o3.csv};

\addplot[only marks, mark=*, mark size=3pt, llamacolor] 
    table[col sep=comma, x=x, y=y]{resources/plots/synth_analysis/task_Llama-4-Maverick-17B-128E-Instruct-FP8.csv};

\addplot[only marks, mark=*, mark size=3pt, claudecolor] 
    table[col sep=comma, x=x, y=y]{resources/plots/synth_analysis/task_claude-sonnet-4-20250514.csv};

\addplot[only marks, mark=*, mark size=3pt, deepseekcolor] 
    table[col sep=comma, x=x, y=y]{resources/plots/synth_analysis/task_deepseek-r1-0528.csv};

\addplot[only marks, mark=square*, mark size=3pt, gpt5nanocolor] 
    table[col sep=comma, x=x, y=y]{resources/plots/synth_analysis/task_gpt-5.2.csv};

\addplot[only marks, mark=square*, mark size=3pt, geminicolor] 
    table[col sep=comma, x=x, y=y]{resources/plots/synth_analysis/task_gemini-3-pro-preview.csv};

\end{axis}

\path let \p1 = (left.west), \p2 = (right.east) in 
    node[anchor=south, draw=black, inner sep=6pt, font=\small,
         minimum width=\x2-\x1-45, align=left] 
    at ($(left.north)!0.32!(right.north)+(0,0.65cm)$) {%
    \raisebox{0.5ex}{\tikz\fill[geminicolor] (0,0) circle (2pt);}~gemini-2.5-pro\hspace{0.8em}%
    \raisebox{0.5ex}{\tikz\fill[geminicolor] (0,0) rectangle (4pt,4pt);}~gemini-3-pro\hspace{0.8em}%
    \raisebox{0.5ex}{\tikz\fill[gpt5nanocolor] (0,0) circle (2pt);}~gpt-5-nano\hspace{0.8em}%
    \raisebox{0.5ex}{\tikz\draw[gpt5minicolor, very thick] (-2pt,-2pt) -- (2pt,2pt) (-2pt,2pt) -- (2pt,-2pt);}~gpt-5-mini\hspace{0.8em}%
    \raisebox{0.5ex}{\tikz\draw[gpt5color, line width=2pt] (-4pt,0) -- (4pt,0);}~gpt-5\hspace{0.8em}%
    \raisebox{0.5ex}{\tikz\fill[gpt5nanocolor] (0,0) rectangle (4pt,4pt);}~gpt-5.2\\[3pt]%
    \raisebox{0.5ex}{\tikz\fill[o4minicolor] (0,0) circle (2pt);}~o4-mini\hspace{0.8em}%
    \raisebox{0.5ex}{\tikz\draw[o3color, very thick] (-3pt,0) -- (3pt,0) (0,-3pt) -- (0,3pt);}~o3\hspace{0.8em}%
    \raisebox{0.5ex}{\tikz\fill[llamacolor] (0,0) circle (2pt);}~Llama-4-Maverick-17B-128E-Instruct-FP8\hspace{0.8em}%
    \raisebox{0.5ex}{\tikz\fill[claudecolor] (0,0) circle (2pt);}~claude-sonnet-4-20250514\\[3pt]
    \raisebox{0.5ex}{\tikz\fill[deepseekcolor] (0,0) circle (2pt);}~deepseek-r1-0528%
};

\end{tikzpicture}
\caption{
    The performance trend for models by type of answer and type of task on \ours-synth. 
}
\label{fig:answer_type}
\end{figure*}

\subsection{\ours-real}

On \ours-real, we divide the questions into ten categories: counting, enumeration, frequency, indexing, player, character, roll type/value, spell level/type (\autoref{tab:oolong_real_single}, \autoref{tab:oolong_real_multi_spells}. \autoref{tab:oolong_real_multi_rolls}). Across models, we find `frequency' to be the hardest and `spell type' the easiest (\autoref{fig:results_by_length_and_type_dnd}). Except for Gemini-2.5-Pro and GPT-5, all models struggle with enumeration and indexing questions. Gemini-2.5 Pro and GPT-5 also make effective use of the player-character mapping provided in the system prompt. 

In  \autoref{fig:results_by_length_and_type_dnd}, we report model performance by question type on \ours-real.

\begin{figure*}
\centering
\input{resources/plots/real_type_windows}

\caption{
    Scores by context window length for various question types in \ours-real. Some questions are only relevant for multi-episode settings.
}
\label{fig:results_by_length_and_type_dnd}
\end{figure*}

\newpage
\clearpage
\section{\ours-real data}
\label{app:dnd_data_rocks}
\subsection{Data quality and curation}

The episode transcripts were first hand-transcribed by members of the fan community, who implemented a multi-step editing process with editorial guidelines. These were then further cleaned by \citet{rameshkumar-bailey-2020-storytelling} to resolve any remaining inconsistencies. We refer the reader to the original paper for more details on the curation process. The stats were manually compiled by volunteers and the team at CritRoleStats. While it’s not possible to completely eliminate sources of potential error, these types of fan-documentation processes are fastidiously annotated and cross-checked by multiple people– a standard of care far higher than that of a paid annotator with no connection to the task or data domain.

\subsection{Challenges of \ours-real}

In the tables below, we highlight various unique challenges presented by \ours-real. \autoref{fig:dnd_example1} covers text that includes in-character, out-of-character and narrative content. Additionally, we show complex discussions surrounding dice rolls (\autoref{fig:dnd_example2}, \autoref{fig:dnd_example4}) and damage calculation (\autoref{fig:dnd_example3}).

\begin{tcolorbox}[colback=gray!10!white, colframe=gray!50!black, boxrule=0.5pt, sharp corners]

\textbf{Matt: }All right. As your smaller friend curls awake, Sam, would you like to describe your character? \\
\textbf{Sam:} Yeah. Um. I am a little goblin girl. (laughter) \\
\textbf{Sam:} I am a goblin. So, you know, the green skin, the green hair, the yellow eyes. And she wears not-great clothes, just like her traveling companion there. She hides in the shadows a lot, because she knows goblins aren't welcome in this part, and that's about it. I mean, she's a little skittish, and right now she's probably stirring awake as well, right? \\
\textbf{Matt:} Well. What's your name? \\
\textbf{Sam:} Oh. Nott the Brave. \\
\textbf{Matt:} So Caleb, as you come to consciousness, you glance over and can see, slowly snoring and rousing at about the same time, Nott's eyes blink open, her slowly groaning face looking over towards you. \\
\textbf{Sam:} (high-pitched Cockney accent) Oh! You're finally awake, I see. Oh yeah, motherfuckers. It's on. You were out for quite some time, there. Rough day, eh? \\
\textbf{Liam:} (light German accent) Not our best day, no.\\
\textbf{Sam: }No, I mean, usually you're so good at everything, but yesterday you were just-- maybe you needed the sleep, is what you needed.

\captionof{figure}{Example of mixed in-character, out-of-character, and narrative text.}
\label{fig:dnd_example1}
\end{tcolorbox}

\begin{tcolorbox}[colback=gray!10!white, colframe=gray!50!black, boxrule=0.5pt, sharp corners]
\textbf{Matt}: The old man's body stops quaking, his flesh now grey and mangled like an ancient tree trunk. He turns, his eyes blood-red and bulging, his lips curled into a horrifying grimace. The two Crown's Guard begin to try and make their way through the panicked crowd, but the people, like a wave of chaos, are keeping them at bay. I need everyone to roll initiative. (yelling) \\
\textbf{Liam}: The miniatures come out, guys! \\
\textbf{Travis}: First map! (yelling) \\
\textbf{Marisha}: I rolled a natural one. \\
\textbf{Sam}: That bodes well. \\
\textbf{Matt}: So you guys, this guy is actually over here. You guys are all up here in the front. \\
\textbf{Laura}: Oh yeah, we were right in the fucking front row. Look at us! We're so cute! \\
\textbf{Taliesin}: Oh my god, it's so pretty. \\
\textbf{Sam}: We're all going to die. \\
\textbf{Liam}: Probably. But then the third campaign begins. \\
\textbf{Laura}: Yay. \\
\textbf{Sam}: I like the tents and donkeys outside. Amazing. \\
\textbf{Marisha}: Look, Taliesin, it's a bardo. Oh my god! \\
\textbf{Taliesin}: No, go for it. Oh boy. \\
\textbf{Matt}: All right, so. 20 to 15? \\
\textbf{Laura}: 21. \\
\textbf{Matt}: All right. \\
\textbf{Laura}: Oh, wait, 22. \\
\textbf{Matt}: Nice. 15 to ten? \\
\textbf{Taliesin}: Ten. \\
\textbf{Sam}: Ten. \\
\textbf{Travis}: 12. \\
\textbf{Ashley}: 13. \\
\textbf{Liam}: 11. \\
\textbf{Matt}: So 13 and then ten and ten? Sorry, 12. So Yasha got 13, then we have Fjord. And then 11. And then we have Nott at ten. And what'd you get? \\
\textbf{Marisha}: Five. Rolled terribly. \\
\captionof{figure}{A complex discussion of dice rolls. While many numbers are mentioned, there are actually seven rolls discussed in this segment: one initiative value for each player.}
\label{fig:dnd_example2}
\end{tcolorbox}

\begin{tcolorbox}[colback=gray!10!white, colframe=gray!50!black, boxrule=0.5pt, sharp corners]
\textbf{Matt}: 16 hits. Go ahead and roll damage. \\
\textbf{Laura}: Yay. 4d6. \\
\textbf{Matt}: Yasha, you're almost on deck. \\
\textbf{Laura}: Ten, 16, 18, and then any attacks in the future-- the next attack has advantage. \\
\textbf{Matt}: How much damage was that? 2d6, 18? \\
\textbf{Sam}: 4d6, you said. \\
\textbf{Laura}: Yeah, it was 4d6 radiant damage. So yeah, it was 18. And then I'm going to run the opposite direction away from him, but I'm going to keep my duplicate up there. \\
\textbf{Matt}: Okay, which, for your duplicate, I should go ahead and grab something. \\
\captionof{figure}{A complex discussion of a damage roll calculation. Many numbers are floated in this discussion; the model must recognize that the discussion refers to a single damage event and the final, resolved damage was 18. }
\label{fig:dnd_example3}
\end{tcolorbox}

\begin{tcolorbox}[colback=gray!10!white, colframe=gray!50!black, boxrule=0.5pt, sharp corners]
\textbf{Sam}: I thought you just said I had advantage! \\
\textbf{Matt}: You had advantage, but you have disadvantage because they're beyond the-- \\
\textbf{Liam}: It cancelled out your advantage. It's just a straight roll and you lost the sneak attack. \\
\textbf{Sam}: Could I run forward and still get the sneak attack? \\
\textbf{Matt}: You could try it. \\
\textbf{Sam}: That's what I will do. Undo. \\
\textbf{Laura}: Roll again, because you might get natural 20! \\
\textbf{Sam}: Now I have to roll for stealth, right? \\
\textbf{Matt}: You've been stealthed as you crept up on this from the last time, so go for it. \\
\textbf{Sam}: I'm rolling to hit? \\
\textbf{Matt}: You already hit on that. \\
\textbf{Laura}: Roll again for advantage, you might get a 20. \\
\textbf{Sam}: Okay. Still 17 plus six. \\
\textbf{Matt}: Now you get a sneak attack, go for it. \\
\captionof{figure}{This discussion highlights a \emph{retcon}: a moment where a dice roll is ``undone'' in the narrative of the game because of a mistake by a player. To tackle, the models needs to understand the narrative around the original roll and correct its count; the original and re-done roll should together count as a single roll.}
\label{fig:dnd_example4}
\end{tcolorbox}

\end{document}